\def\year{2021}\relax
\documentclass[letterpaper]{article} % DO NOT CHANGE THIS
\usepackage{aaai21}  % DO NOT CHANGE THIS
\usepackage{times}  % DO NOT CHANGE THIS
\usepackage{helvet} % DO NOT CHANGE THIS
\usepackage{courier}  % DO NOT CHANGE THIS
\usepackage[hyphens]{url}  % DO NOT CHANGE THIS
\usepackage{graphicx} % DO NOT CHANGE THIS
\usepackage{natbib}  % DO NOT CHANGE THIS AND DO NOT ADD ANY OPTIONS TO IT
\usepackage{caption} % DO NOT CHANGE THIS AND DO NOT ADD ANY OPTIONS TO IT
\usepackage{subfig}
\usepackage{amssymb}
\usepackage{amsthm}
\usepackage{lmodern, amsmath}
\usepackage{algorithmic}
\usepackage[ruled,vlined]{algorithm2e}
\newcommand{\option}{\text{O}}
\newtheorem{theorem}{Theorem}

\makeatletter
\gdef\copyright@on{}
\makeatother
\title{Diversity Enriched Option-Critic}

\author{
    Anand Kamat \textsuperscript{\rm 1,2},
    Doina Precup \textsuperscript{\rm 1,2,3}\\
}
\affiliations{
    \textsuperscript{\rm 1} McGill University  
    \textsuperscript{\rm 2} Mila     
    \textsuperscript{\rm 3}Google Deepmind \\
    anand.kamat@mail.mcgill.ca, dprecup@cs.mcgill.ca
}

\begin{document}

\maketitle

\begin{abstract}
Temporal abstraction allows reinforcement learning agents to represent knowledge and develop strategies over different temporal scales. The option-critic framework has been demonstrated to learn temporally extended actions, represented as options, end-to-end in a model-free setting. However, feasibility of option-critic remains limited due to two major challenges, multiple options adopting very similar behavior, or a shrinking set of task relevant options. These occurrences not only void the need for temporal abstraction, they also suppress performance. In this paper, we tackle these problems by learning a \textit{diverse set of options}. We introduce an information-theoretic intrinsic reward, which augments the task reward, as well as a novel termination objective, in order to encourage behavioral diversity in the option set. We show empirically that our proposed method is capable of learning options end-to-end on several discrete and continuous control tasks, outperforms option-critic by a wide margin. Furthermore, we show that our approach sustainably generates robust, reusable, reliable and interpretable options, in contrast to option-critic.
\end{abstract}

\section{Introduction}
\label{introduction}
% Temporal abstraction has spawn a lot of interest in recent years. 
%Humans have a remarkable ability to overcome arduous tasks by learning, planning and representing knowledge hierarchically. 
The ability of reinforcement learning (RL) agents to solve very large problems efficiently depends on building and leveraging knowledge that can be re-used in many circumstances. One type of such knowledge comes in the form of 
options  \cite{Sutton:1999:MSF:319103.319108, Precup2000TemporalAI}, temporally extended actions that can be viewed as specialized skills which can improve learning and planning efficiency \cite{Precup2000TemporalAI,TRIO}. The option-critic framework \cite{bacon2017option} proposed a formulation to learn option policies as well as the termination conditions end-to-end, through gradient descent, just from the agent's experience and rewards. However, this can lead to the option set collapsing in various ways \cite{bacon2017option,termination-critic2019}, for example options becoming primitive actions, one option learning to solve the entire task and dominating the others, or several options becoming very similar. These degeneracies not only negatively impact the agent's ability to re-use the learned option set in new situations, they often hurt performance. Furthermore, learning options in a model-free setting is often accompanied with increased sample and computational complexity over primitive action policies, without the desired performance improvements. This raises the fundamental question of why temporal abstraction is needed, especially when a primitive action policy achieves comparable results. 
There have been attempts to tackle the problem of options collapsing onto multiple copies of the optimal policy \cite{bacon2017option,termination-critic2019} as well as ensuring that options do not shrink too much over time \cite{deliiberationcost}. However, finding a solution that can easily generalize over a wide range of tasks with minimal human supervision is still an ongoing challenge. In this paper, we tackle the problem by constructing a \emph{diverse set of options}, which can be beneficial to increase exploration as well as for robustness in learning challenging tasks \cite{gregor2016variational,diaynpaper-2018,termination-critic2019}. 
A common approach for encouraging diversity in a policy is entropy regularization \cite{Williams1991FunctionOU, Mnih2016AsynchronousMF}, but it does not capture the idea of the set of options itself containing skills that differ from each other. Unlike in the case of primitive action policies where each action is significantly distinct, options often learn similar skills reducing the effectiveness of entropy regularization. To address this issue, we use intrinsic motivation. Augmenting the standard maximum reward objective with auxiliary bonus have encouraging results in promoting good exploration and performance \cite{ng1999policy, singh2010intrinsically, count_1}. We introduce an auxiliary reward which, when combined with the task reward, encourages diversity in the policies of the options. We empirically show how this diversity can help options explore better on several standard continuous control tasks.
% Degeneration of options
We then focus on option termination. The termination objective used in option-critic \cite{bacon2017option} increases the likelihood of an option to terminate if the value of the current option is sub-optimal with respect to another. Though logical, this objective tends to suppress the worse option quickly without adequate exploration, often resulting in a single option dominating over the entire task \cite{bacon2017option, deliiberationcost, termination-critic2019}. To overcome this, we also propose a novel termination objective which produces diverse, robust and task relevant options. Our approach suggests that instead of having options compete for selection, adequate exploration of available options should be encouraged so long as they exhibit diverse behavior. Upon testing this new objective quantitatively and qualitatively in a classic tabular setting as well as several standard discrete and continuous control tasks, we demonstrate that our approach achieves a new state-of-the-art performance. Furthermore, our approach demonstrates significant improvements in robustness, interpretibility and reusability of specialized options in transfer learning. 
% End of introduction
\begin{figure*}[!ht]
    \centering
    \subfloat[HalfCheetah-v2]{\includegraphics[scale=0.12]{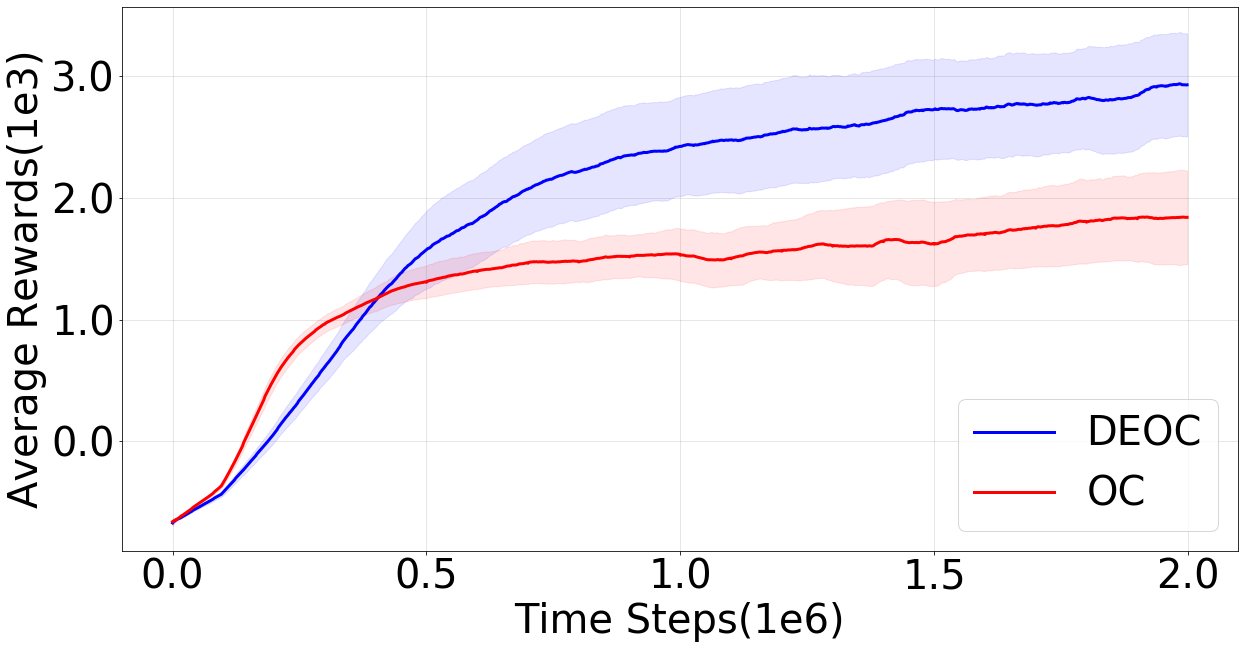}  \label{Fig_DEOCvsPPOC_HalfCheetah}}
    \subfloat[Hopper-v2]{\includegraphics[scale=0.12]{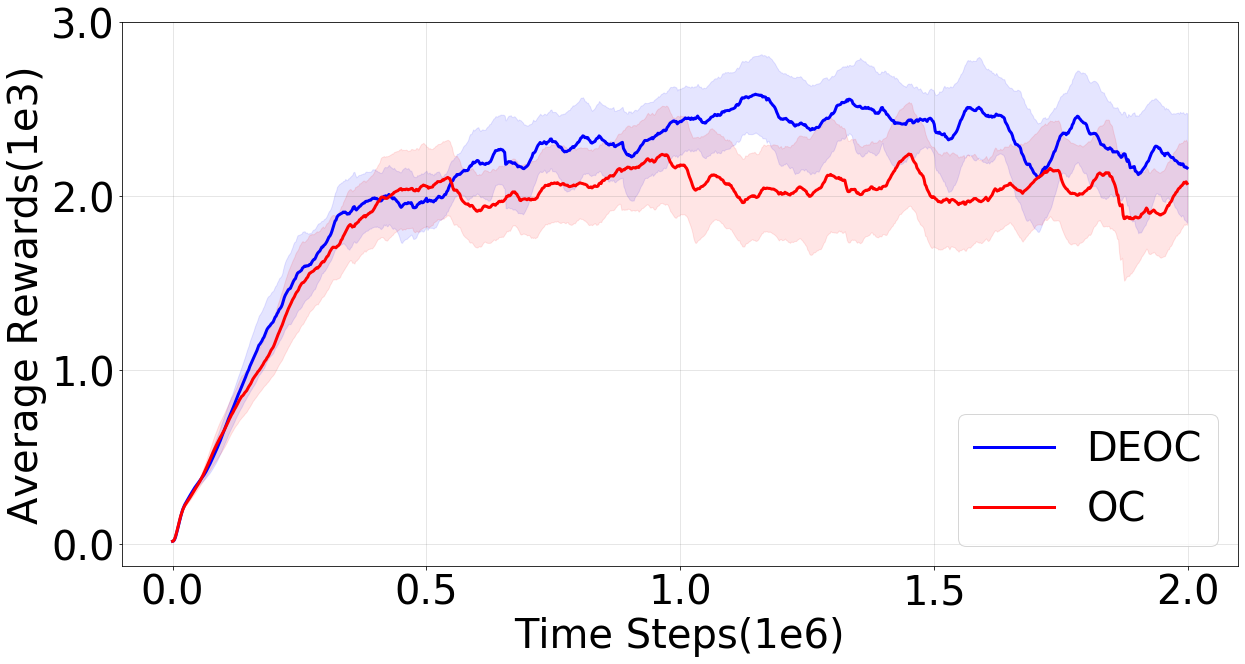}\label{Fig_DEOCvsPPOC_Hopper}}
    \subfloat[Walker2d-v2]{\includegraphics[scale=0.12]{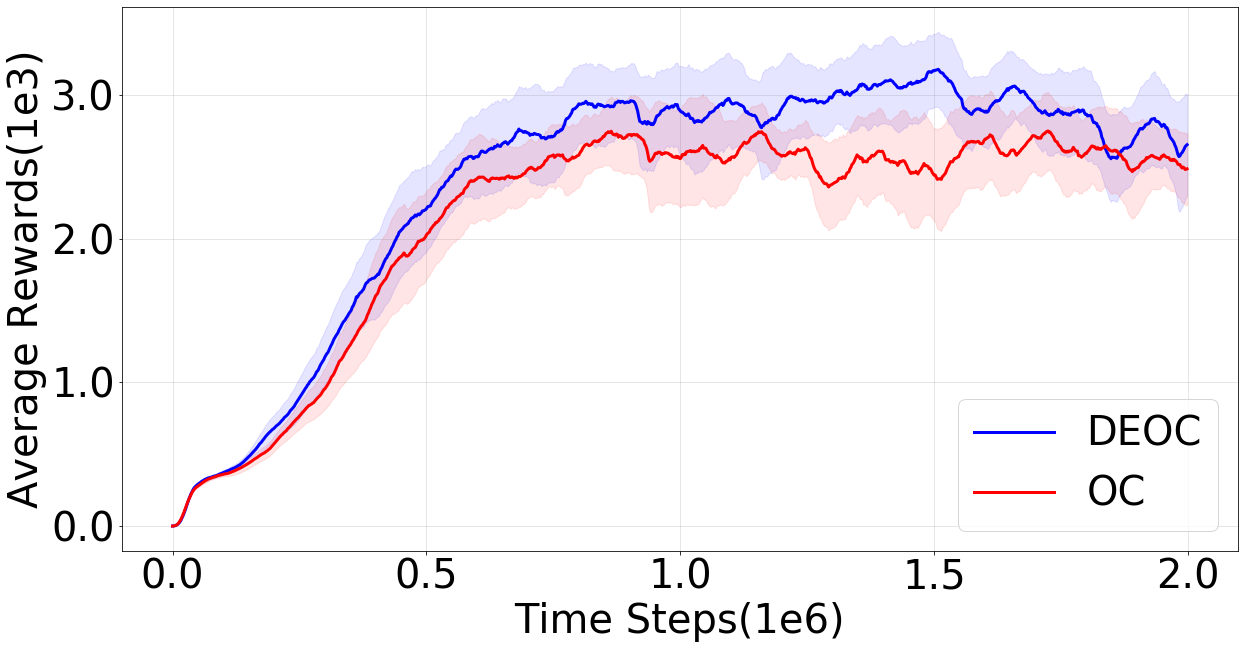}\label{Fig_DEOCvsPPOC_Walker}}
    \caption{\textbf{Diversity-Enriched Option-Critic (DEOC) compared against Option-Critic (OC)}. Each plot is averaged over 20 independent runs.
    }
    \label{Fig_DEOCvsPPOC}
\end{figure*}
\section{Background}
% Reinforcement Learning
In RL, an agent interacts with an environment typically assumed to be a Markov Decision Process (MDP) $\mathcal{M} = (\mathcal{S}, \mathcal{A},\gamma, r, \mathcal{P})$ where $\mathcal{S}$ is the set of states,  $\mathcal{A}$ is the action set, $\gamma \in [0,1)$ is the discount factor,  $r:\mathcal{S} \times \mathcal{A} \rightarrow \mathbb{R}$ is the reward function and $\mathcal{P} :\mathcal{ S} \times \mathcal{A} \times \mathcal{S} \rightarrow [0, 1]$ is the transition dynamics. A policy $\pi$ is a probabilistic distribution over actions conditioned on states, $\pi: \mathcal{S} \times \mathcal{A} \rightarrow [0, 1]$ . The value function of a policy $\pi$ is  the expected discounted return $V_{\pi}(s) = \mathbb{E}_{\pi} \Big[ \sum_{t=0}^{\infty} \gamma^{t}r_{t+1}|s_{0} = s \Big] $. 
% Policy Gradient algorithms
Policy gradient methods aim to find a good policy by optimizing the expected return over a given family of parameterized stochastic policies $\pi_{\theta}$. Policy improvement is carried out by performing stochastic gradient ascent over the policy parameters. 
Techniques for defining useful abstractions through hierarchy have sparked a lot of interest \cite{parr1998reinforcement, dietterich2000hierarchical, Sutton:1999:MSF:319103.319108, Precup2000TemporalAI, mcgovern2001automatic, Stolle_learningoptions, Vezhnevets2017FeUdalNF}.
% Options Framework
We use the options framework \cite{Sutton:1999:MSF:319103.319108,Precup2000TemporalAI}, which formalizes temporal abstraction by representing knowledge, learning and planning over different temporal scales. An option \textit{o} is a temporally extended action represented as a triple ($\mathcal{I_{\textit{o}}, \pi_{\textit{o}}, \beta_{\textit{o}} }$) where $\mathcal{I_{\textit{o}} \subseteq S }$ is an initiation set, $\pi_{\textit{o}}$ is an intra-option policy and $\beta_{\textit{o}}$: $\mathcal{S} \rightarrow [0,1]$ is a termination function. 
The policy over options $\pi_{\Omega}$, selects an option from those available  at a given state and executes it until termination.  Upon termination, $\pi_{\Omega}$ selects a new option and this process is repeated. The option-critic architecture \cite{bacon2017option} is a gradient-based method for learning options end-to-end without explicitly providing any sub-goals, by updating the  parameters of intra-option policies $(\theta_{\pi})$ and terminations ($\theta_{\beta}$).
The termination gradient in option-critic~\cite{bacon2017option} states that at a state, if the value of an option is sub-optimal compared to the value of the policy over options, the likelihood of its termination should be increased. as follows:
\begin{equation} \label{eq_terminationgradient}
    \frac{ \partial L(\theta) }{ \partial \theta_{\beta} }  =  \mathbb{E} \bigg[- \frac{ \partial \beta (S_{t},O_{t}) }{ \partial \theta_{\beta} } A(S_{t},O_{t}) \bigg]
\end{equation}
where $ A(S_{t},O_{t}) = Q_{\pi}(S_{t},O_{t}) - V_{\pi}(S_{t}) $ is the termination advantage function.
% Works in temporal abstraction and related literature

Since primitive actions are sufficient for learning any MDP, options often degenerate  \cite{bacon2017option}. Several techniques have been proposed to tackle this problem \cite{bacon2017option,deliiberationcost,termination-critic2019}. 
In the next two sections, we outline the main idea for our approach: encouraging diversity in option policies and encouraging the options to terminate in diverse locations.
% ______________________________________________________________________
\section{Encouraging Diversity While Learning}\label{IntrinsicReward_Section}

A good reward function  can capture more than just information required to perform the task. In this section, we highlight the importance of diversity in options and an approach to achieve it using intrinsic motivation. We design a pseudo-reward signal complementing the task reward in order to encourage diversity.
% Our approach shares a similar idea as \citet{diaynpaper-2018} for designing a pseudo reward signal that encourages diversity in options. 
While most relevant literature on learning diverse options \cite{gregor2016variational,diaynpaper-2018,termination-critic2019} use states to distinguish and specialize options, we instead look directly at an option's behavior to assess its diversity. This idea is well suited when all options are available everywhere, when the state information is imperfect (for example, because the latent representation of the state is still being learned), and when the agent aims to transfer knowledge across tasks. This approach allows the agent to effectively reuse specialized options \cite{bacon2017option}. For example, an option specialized in leaping over hurdles can be reused if and when the agent encounters a hurdle anywhere in its trajectory. We study reusability and transfer characteristics of options in Section \ref{section_transfer_tasks}. 
For simplicity of exposition, we use two options in our notation in this chapter; however the approach can be easily extended to any number of options. An empirical study with varying number of options are presented in Appendix \ref{app_variable_options}.
We construct our pseudo reward function using concepts from information theory. Maximizing the entropy of a policy prevents the policy from quickly falling into a local optimum and has been shown to have substantial improvements in exploration and robustness \cite{Williams1991FunctionOU, Mnih2016AsynchronousMF, haarnoja2018soft}. We maximize the entropy $\mathcal{H}( A^{\pi_{\text{\option}_{1}}} \mid  S)$ and $\mathcal{H}( A^{\pi_{\text{\option}_{2}}} \mid  S)$ where $ \mathcal{H} $ is the Shannon entropy computed with base \textit{e} and $ A $ represents respective action distributions. Since we want different options behave differently from each other at a given state, we maximize the divergence between their action distributions $\mathcal{H}(A^{\pi_{\text{\option}_{1}}}; A^{\pi_{\text{\option}_{2}}}\mid S)$.  This aligns with our motivation that skill discrimination should rely on actions. 
% Since we want different options behave differently from each other at a given state, we minimize the mutual information ($I$) between their action distributions ($ A $) at that state i.e. $I( A^{\pi_{\text{\option}_{1}}} ;  A^{\pi_{\text{\option}_{2}}} \mid  S)$. This aligns with our motivation that skill discrimination should rely on actions. 
% Entropy regularization is a commonly used strategy to increase stochasticity of a policy thereby preventing the policy from quickly falling into a local optimum.
% We maximize the entropy $\mathcal{H}( A^{\pi_{\text{\option}_{1}}} \mid  S)$ and $\mathcal{H}( A^{\pi_{\text{\option}_{2}}} \mid  S)$ where $ \mathcal{H} $ is the Shannon entropy. %Both $I$ and $ \mathcal{H} $ are computed with base \textit{e}. 
Lastly, we seek to maximize the stochasticity of the policy over options $\mathcal{H}( O^{\pi_{\Omega}} \mid  S)$ to explore all available options at $S$. Combining all the above terms, we get the following pseudo reward  $ \mathcal{R}_{bonus} $: 
\begin{align}\label{eq_pseudoreward}
\mathcal{R}_{bonus} &=  \mathcal{H}(A^{\pi_{\text{\option}_{1}}} \mid S) + \mathcal{H}(A^{\pi_{\text{\option}_{2}}} \mid S) \nonumber \\
&+ \mathcal{H}( O^{\pi_{\Omega}} \mid S) + \mathcal{H}(A^{\pi_{\text{\option}_{1}}}; A^{\pi_{\text{\option}_{2}}}\mid S)
\end{align}
The first three terms in Eq. \eqref{eq_pseudoreward} seeks to increase the stochasticity of the policies and the fourth term encourages overall diversity in options. Since we use entropy regularization for policy updates in all our implementations as well as baseline experiments, we only use $\mathcal{H}(A^{\pi_{\text{\option}_{1}}}; A^{\pi_{\text{\option}_{2}}}\mid S)$ as our pseudo reward, $\mathcal{R}_{bonus}$, highlighting the significance of diversity in the option set.

We incorporate this objective within the standard RL framework by augmenting the reward function to include the pseudo reward bonus from Eq. \eqref{eq_pseudoreward}:
\begin{equation}\label{Eq_reward_augmentation}
    \mathcal{R}_{aug}(S_{t},A_{t}) = (1 - \tau)R(S_{t},A_{t}) + \tau \mathcal{R}_{bonus}(S_{t}) 
\end{equation}
where  $ \tau $ is a hyper-parameter  which controls relative importance of the diversity term against the reward. The proposed reward augmentation yields the maximum diversity objective. The standard RL objective can be recovered in the limit as $\tau \rightarrow 0$.
To demonstrate the benefits of maximizing diversity through augmenting the reward, we test our algorithm, Diversity-Enriched Option-Critic (DEOC), against Option-Critic (OC) in classic Mujoco \cite{todorov2012mujoco} environments. 
% We use the PPOC codebase \cite{baselines,Schulman2017ProximalPO,Klissarov2017LearningsOE} for our experiments.
We use the same hyper-parameter settings across all 20 seeds in all our experiments throughout the paper to test stability. Fig. \ref{Fig_DEOCvsPPOC} shows that encouraging diversity improves sample efficiency as well as performance. 
% The most significant impact was noticed in HalfCheeetah-v2 where a good exploration strategy plays a huge impact. Unlike OC which learns a type of gait where the agent flips over and slides on its back, DEOC almost always manages to run upright thereby showcasing much better results. 
Details regarding implementation and choices for the underlying algorithm, PPO \cite{Schulman2017ProximalPO}, are provided in Appendix \ref{App_Nonlinearcase}.
% \begin{figure}
%     \centering
%     \subfloat[Diversity Enriched Option-Critic (DEOC)]{\includegraphics[scale=0.12]{Figures/cheetah_upright.png}  \label{Fig_uprightcheetah}}
%     \subfloat[Option-Critic (OC)]{\includegraphics[scale=0.12]{Figures/cheetah_flipped.png} \label{Fig_flippedcheetah}} \\
%     \caption{\textbf{Illustrations showing the gaits learned by DEOC and Option-Critic (OC).} Unlike OC where the agent flips over and slides on its back, DEOC almost always learns to run upright.}
%     \label{Fig_cheetah_flipped_and_upright}
% \end{figure}
\section{Encouraging Diversity in Termination}\label{TerminationSection}
In Section \ref{IntrinsicReward_Section}, we empirically demonstrate that encouraging diversity in option policies improves exploration and performance of option-critic. However, unlike primitive action policies where all actions are available at every step, options execute for variable time steps until a termination condition is met, during which, all other options remain dormant. Due to this, the maximum entropy objective fails to be as effective with options as with primitive action policies. Although having options terminate at every time step may solve this problem, it renders the use of options moot.
Additionally, option-critic's termination function solely validates the best option, suppressing other potentially viable options which may also lead to near-optimum behavior. As a consequence, at a given state, only the best current option gets updated, eventually leading to a single option dominating the entire task.
% At a given state, the `worse' option can be quickly suppressed without adequate exploration, while the best current option keeps improving. Eventually, this leads to a single option dominating the entire task. 
Noise in value estimates or state representations may also cause an option to terminate and consequently lead to the selection of a sub-optimal option. Selecting a sub-optimal option around ``vulnerable'' states can be catastrophic and also severely hurt performance. In our case, despite $ \mathcal{R}_{bonus}(s,a)$ encouraging diverse options, option-critic's termination function prevents exploiting this diversity due to inadequate exploration of all relevant options. 
% computing  $ \mathcal{R}_{bonus}(s,a)  $ while one of the options is never selected fails the purpose of generating diverse options.  
% Furthermore, since primitive actions are sufficient for solving any MDP, options often default into primitive actions. 
% Unfortunately, the bonus reward $ \mathcal{R}_{bonus}(s,a)  $, does not take into account the option's relevance at any state. Computing  $ \mathcal{R}_{bonus}(s,a)  $ while one of the options is never selected fails the purpose of generating diverse options. 
We tackle these problems by encouraging all options available at a given state to be explored, so long as they exhibit diverse behavior.\\
In this section we present a novel termination objective which no longer aims to maximize the expected discounted returns, but focuses on the option's behavior and identifying states where options are distinct, while still being relevant to the task. 
We build our objective function to satisfy the following two conditions:
\begin{itemize}
\item \textbf{Options should terminate in states where the available options are diverse.} 
% The $\mathcal{H}( O^{\pi_{\Omega}} \mid S)$ regularizer can now fairly select and exploit the diverse options. 
In the classic four-rooms task \cite{Sutton:1999:MSF:319103.319108}, such states would be the hallways.
% around which different options can easily adopt different navigation strategies such as entering the room or turning back, depending on the location of the goal state. 
Terminations localized around hallways have shown significant improvements in performance in the transfer setting \cite{bacon2017option}.
    % The termination objective needs to be decoupled from the standard expected return objective} which most reinforcement learning algorithms aim to maximize. Such an objective (Eq \eqref{eq_terminationgradient}) has shown to prefer the option that gets slightly better than the rest at a given state, and fails to select and explore other options. Instead of competing for dominance, available options should be explored as long as they exhibit different strategies.  
\item \textbf{The diversity metric used in the termination objective should capture the diversity relative to other states in the sampled trajectories.} This prevents options terminating at every step while exploiting diversity effectively for exploration and stability.

% Diverse options can be best exploited for exploration and stability around these states.
% We wish terminations to focus on states where options are most diverse, relative to other observed states. This makes sense as diversity in options can be better exploited for exploration and stability around these states.
\end{itemize}
\begin{algorithm}[!t]
   \caption{Termination-DEOC (TDEOC) algorithm with tabular intra-option Q-Learning}
   \label{deoc_algo}
\begin{algorithmic}
    % \STATE Initialize policy over options ($\pi_{\Omega} $)
    % \STATE Initialize intra-option policy ($\pi_{o} $)
    % \STATE Initialize termination function ($\beta_{o} $)
    \STATE Initialize $\pi_{\Omega} $, $\pi{o} $ and $\beta_{o} $
    % \STATE $ s_{t} \leftarrow s_{0} $
    \STATE Choose $ o_{t} $ according to  $ \pi_{\Omega} (o_{t}|s_{0}) $ 
    \REPEAT
        % \STATE Choose $ a_{t} $ according to  $ \pi_{o} (a_{t}|s_{t}) $
        \STATE Act $ a_{t} \sim \pi_{o} (a_{t}|s_{t}) $ and observe $ s_{t+1} $ and $ r_{t} $
        % \STATE Compute $ r_{bonus}(s_{t}) $
        \STATE  $ r'_{t} =  (1 - \tau)r_{t} + \tau \, r_{bonus}(s_{t}) $ 
        \IF{$ o_{t}$ terminates in  $s_{t+1} $ }
            \STATE Choose new $o _{t+1} $ according to $ \pi_{\Omega} (\cdot | s_{t+1}) $
        \ELSE
            \STATE $o_{t+1} = o_{t}$
        \ENDIF
        % \STATE Calculate the moving average and standard deviation of  $r_{bonus}(s_{t})$    
        \STATE $\textit{D}(s_{t}) \leftarrow $ Standardized samples of $r_{bonus}(s_{t})$.
        % \STATE \textbf{Options Evaluation:}
        \STATE $\delta \leftarrow$ $r'_{t}$ - $Q_{U}(s_{t}, o_{t}, a_{t})$
        \STATE $\delta \leftarrow$ $\delta$ $+$ $\gamma$(1 $-$ $\beta_{o_{t}} (s_{t+1}))Q_{\Omega}(s_{t+1},o_{t})$ + \\ 
         $\: \: \:$ $\gamma \beta_{o_{t}} (s_{t+1}) max_{o_{t+1}} Q_{\Omega}(s_{t+1},o_{t+1})$ \\
        $Q_{U}(s_{t}, o_{t}, a_{t}) \leftarrow $ $Q_{U}(s_{t}, o_{t}, a_{t}) + \alpha \delta$  \\
        % \STATE \textbf{Options Improvement:} \\
        \STATE $\theta_{\pi} \leftarrow \theta_{\pi} + \alpha_{\theta_{\pi}} \frac{\partial log \pi_{o_{t}}(a_{t} | s_{t})}{\partial \theta} Q_{U}(s_{t}, o_{t}, a_{t}) $
        \STATE $\theta_{\beta} \leftarrow \theta_{\beta} +  \alpha_{\theta_{\beta}} \frac{\partial \beta_{o_{t}}(s_{t+1})}{\partial \nu}$ $\mathcal{D}(s_{t+1})$
    \UNTIL{$s_{t+1}$ is terminal}
        
\end{algorithmic}
\end{algorithm}
The termination objective we maximize becomes: 
\begin{equation} \label{eq_deocobjective}
     L(\theta_{\beta})  =  \mathbb{E} \big[\beta(S_{t},O_{t})   \mathcal{D}(S_{t})  \big]
\end{equation}
The term $ \mathcal{D}(S_{t}) $ indicates the relative diversity of options at a given state. We compute $ \mathcal{D}(S_{t}) $ by standardizing (with a mean $ \mu=0 $, and standard deviation $ \sigma=1 $), the samples of $ \mathcal{R}_{bonus}(S_{t})$ defined in Eq. \eqref{eq_pseudoreward}, collected in the buffer. 
% To compute $ \mathcal{D}(S_{t}) $, we use samples of the reward ( $ \mathcal{R}_{bonus}(S_{t}) $ ) defined in Eq. \eqref{eq_pseudoreward} which is a measure of how diverse options are at a given state.  However, $ \mathcal{R}_{bonus}(S_{t}) $  is a positive term, which would consequently always increase the termination likelihood for any state. We mitigate this by standardizing (with a mean $ \mu=0 $, and standard deviation $ \sigma=1 $), the $ \mathcal{R}_{bonus}(S_{t}) $ samples collected in the buffer, to obtain $ \mathcal{D}(S_{t}) $. 
\begin{equation} \label{eq_standardize}
     \mathcal{D}(S_{t}) = \frac{\mathcal{R}_{bonus}(S_{t}) - \mu_{\mathcal{R}_{bonus}}}{ \sigma_{\mathcal{R}_{bonus}}}
\end{equation}
% \begin{figure} [!ht]
%     \centering
%     \subfloat[Termination-DEOC (TDEOC)]{\includegraphics[scale=0.21]{Figures/Tabular_TDEOC_terminations-nips.png}  \label{Fig_DEOC_termination_plot}} \\
%     \subfloat[Option-Critic (OC)]{\includegraphics[scale=0.21]{Figures/Tabular_Vanilla_Terminations.png} \label{Fig_OC_termination_plot}} \\
%     \caption{\textbf{Visualization of Terminations for different options} after 1000 episodes. Darker colors correspond to higher termination likelihood. Both TDEOC and OC show higher terminations around hallways.}
%     \label{Fig_Termination_plots}
% \end{figure} 
Our approach solves the issue of constant termination at all states, and  the updates are scaled appropriately relative to the diversity values of other states in the buffer. Terminating while options are most diverse encourages both options to be selected fairly and explored by the policy over options. 
\begin{theorem}\label{terminationtheorem}
Given 
% the gradient-based option-critic algorithm \cite{bacon2017option}, 
a set of Markov options $\Omega$ each with a stochastic termination function defined by Eq. \eqref{eq_deocobjective} and stochastic intra-option policies, with $|\Omega|<\infty$ and $|\mathcal{A}|<\infty$, repeated application of policy-options evaluation and improvement \cite{bacon2017option, Bacon2013phdthesis} yields convergence to a locally optimum solution. 

Proof. See Appendix \ref{app_proofterminationtheorem}.
\end{theorem}

Note that as with $ \mathcal{R}_{bonus}(S_{t}) $, $ \mathcal{D}(S_{t}) $ is independent of the termination parameters. An added advantage of using relative diversity is the agent's ability to respond to events or obstacles in its trajectory.
% An example of such an event could be the presence of a hurdle in a locomotion task.
Such events characterize some of the most sensitive  states in the environment. 
The relative diversity $ \mathcal{D}(S_{t}) $ in our objective is capable of identifying such states, causing both options to collectively explore and learn the event. We study transfer characteristics further in Section \ref{section_transfer_tasks}. 
% ------------------------------------------------------------
\section{Experiments} \label{section_TDEOC_experiments}
We evaluate the effects of the new termination objective on several tasks, to test its performance and stability. The pseudo-code of the algorithm, Termination-DEOC (TDEOC), is presented in Algorithm \ref{deoc_algo}. Implementation details are provided in Appendix \ref{App_Implementation_details}. \par
\begin{figure} [!ht]
    \centering
    \subfloat[Termination-DEOC (TDEOC)]{\includegraphics[scale=0.14]{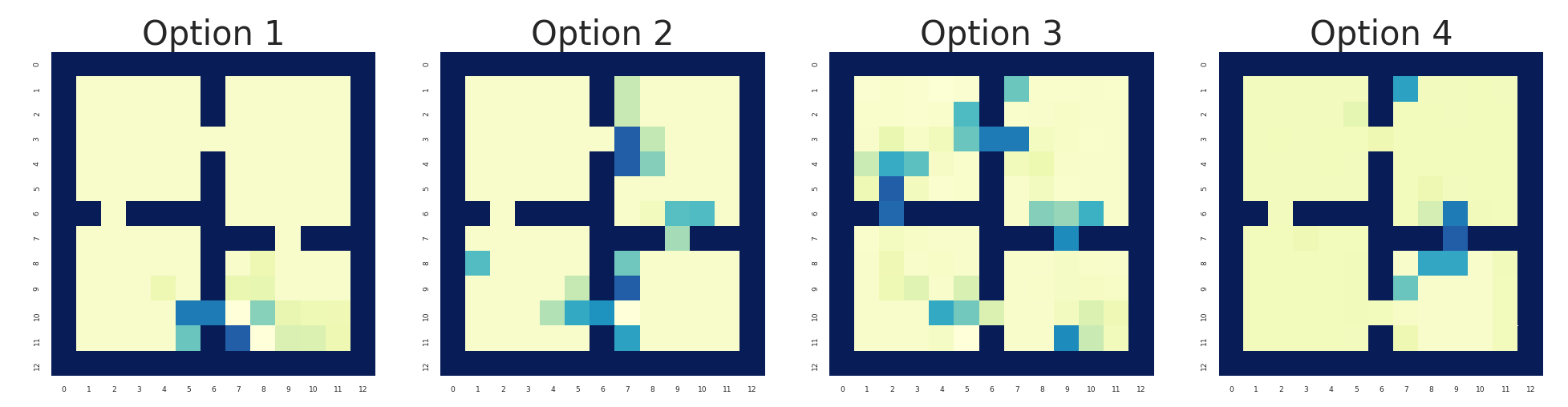}  \label{Fig_DEOC_termination_plot}} \\
    \subfloat[Option-Critic (OC)]{\includegraphics[scale=0.21]{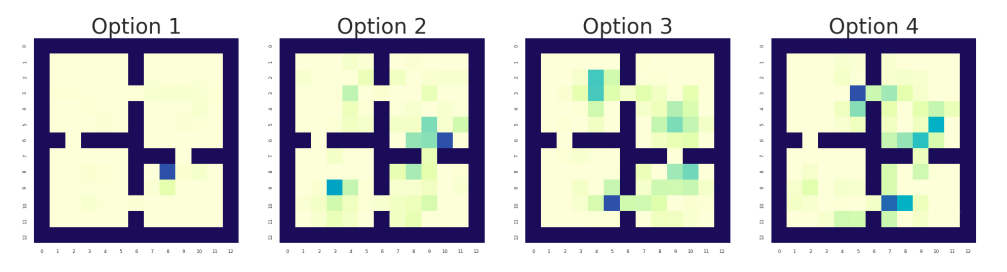} \label{Fig_OC_termination_plot}} \\
    \caption{\textbf{Visualization of Terminations for different options} after 1000 episodes. Darker colors correspond to higher termination likelihood. Both TDEOC and OC show higher terminations around hallways.}
    \label{Fig_Termination_plots}
\end{figure} 
\begin{figure} [!ht]
    \centering
    \subfloat[Termination-DEOC (TDEOC) VS OC]{\includegraphics[scale=0.175]{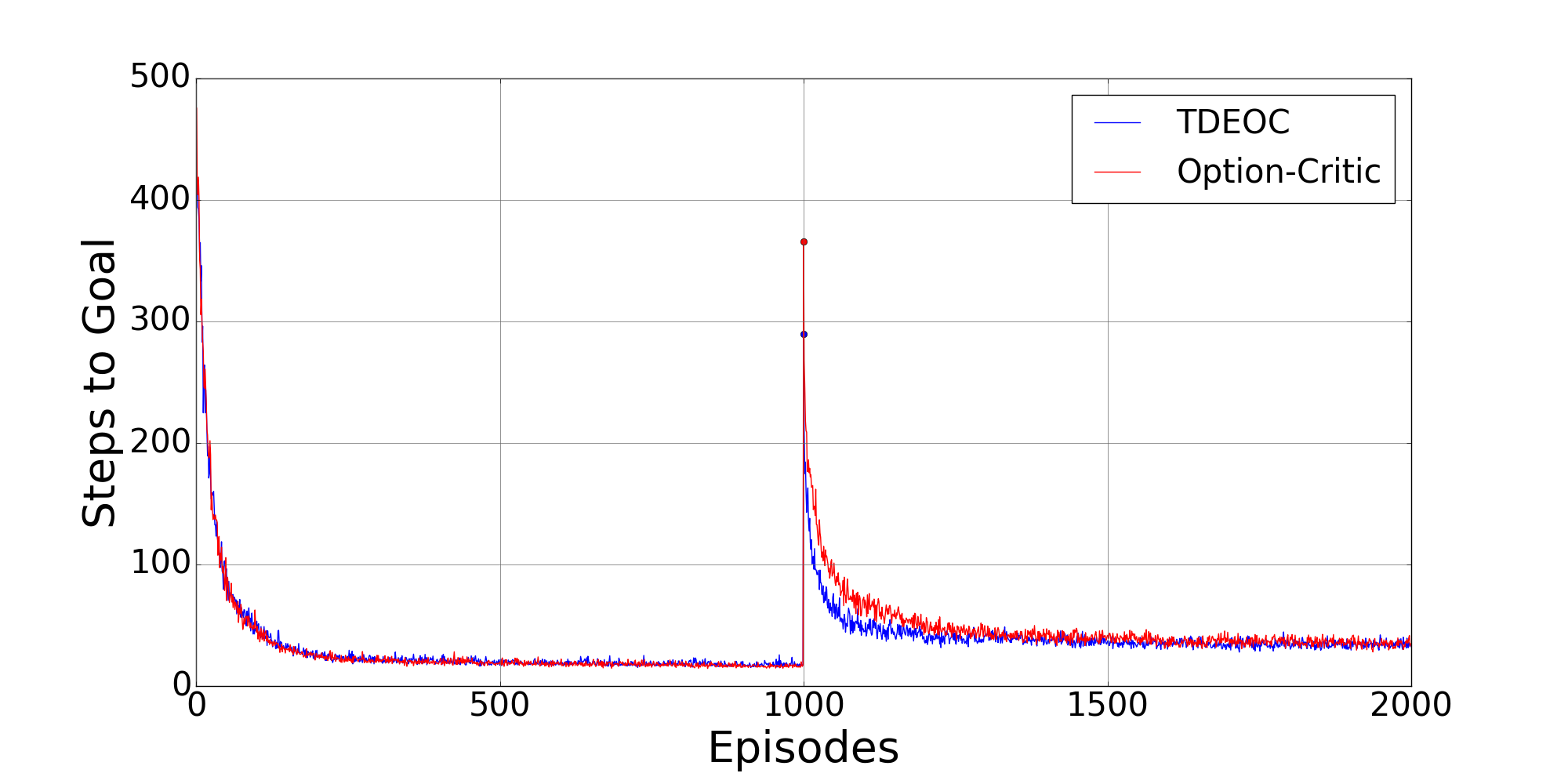} \label{Fig_OC_termination_plot}} \\
    \caption{\textbf{Four-rooms transfer experiment with four options}. After 1000 episodes, the goal state, is moved from the east hallway to a random location in the south east room. TDEOC recovers faster than OC with a difference of almost 70 steps when the task is changed. Each line is averaged over 300 runs.
    }
    \label{Fig_Termination_Fourrooms_plots}
\end{figure}
\begin{figure*}[!ht]
    \centering
     \textsc{\textbf{Empirical Performance}}\\
    \subfloat[Humanoid-v2]{\includegraphics[scale=0.12]{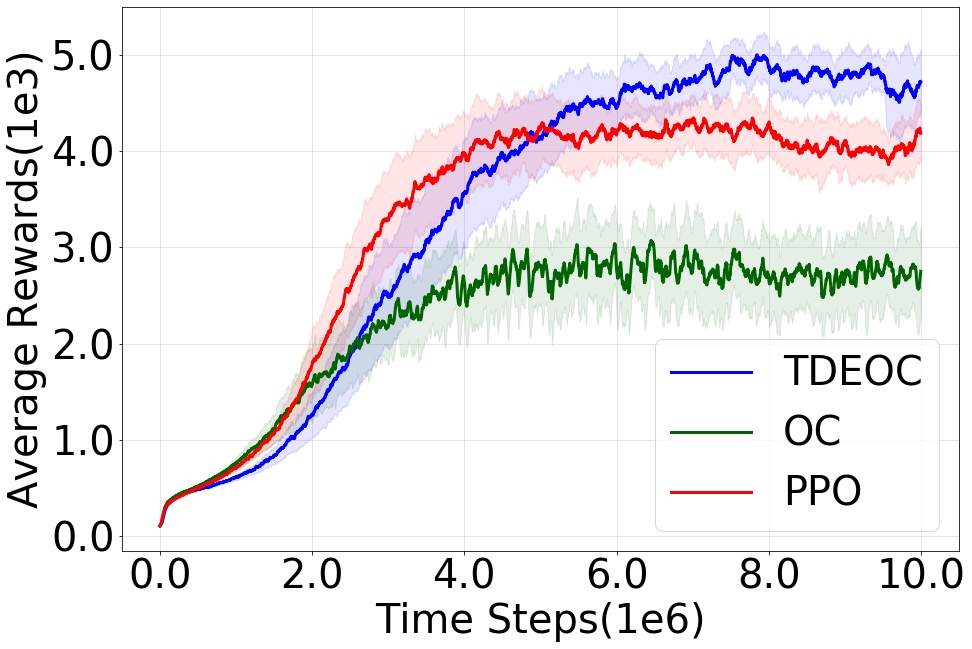} \label{Fig_Humanoid_results}} 
    \subfloat[HalfCheetah-v2]{\includegraphics[scale=0.12]{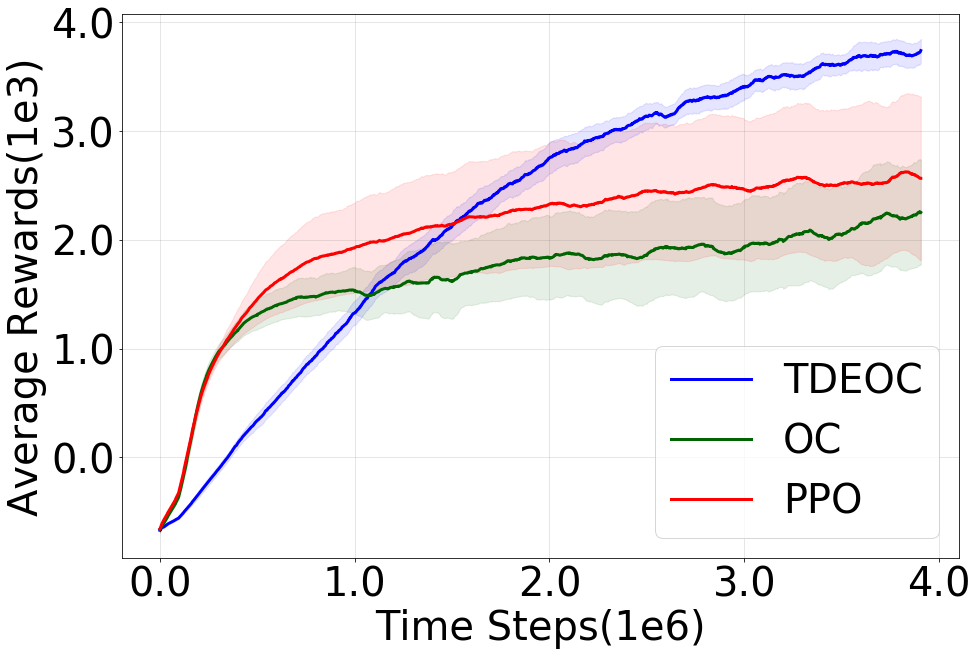}  \label{Fig_Halfcheetah_results}}
    \subfloat[Sidewalk (Discrete)]{\includegraphics[scale=0.12]{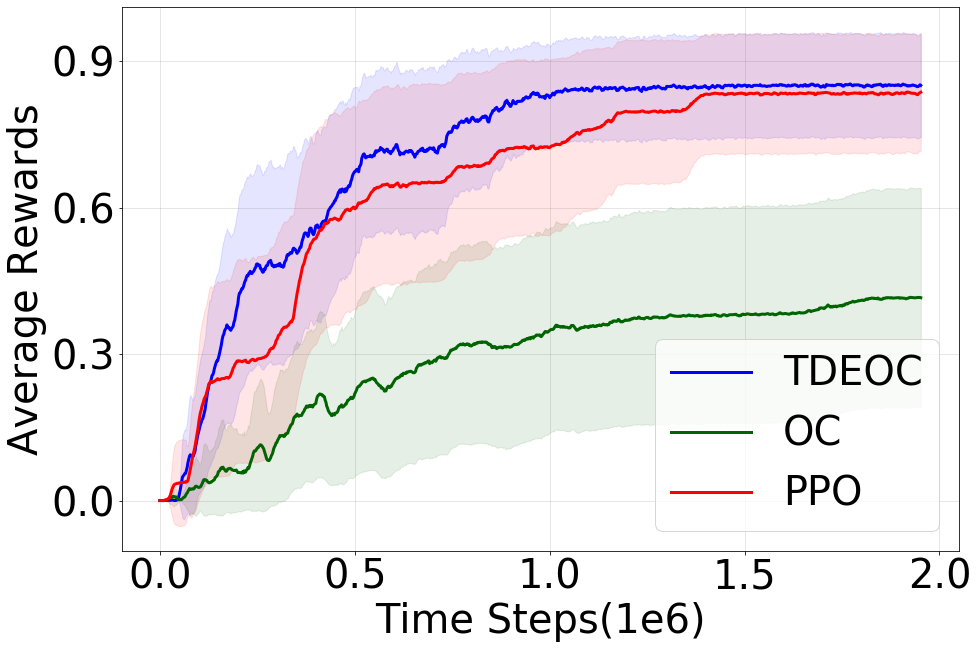} \label{Fig_Sidewalk_results}}\\
    \subfloat[Ant-v2]{\includegraphics[scale=0.12]{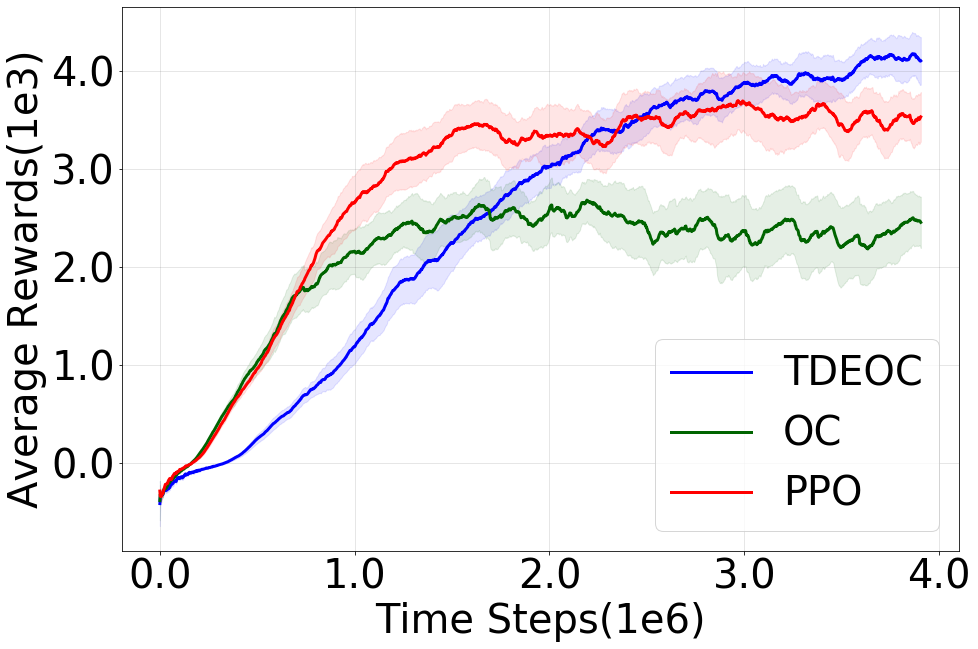} \label{Fig_Ant_results}}
    \subfloat[Walker2d-v2]{\includegraphics[scale=0.12]{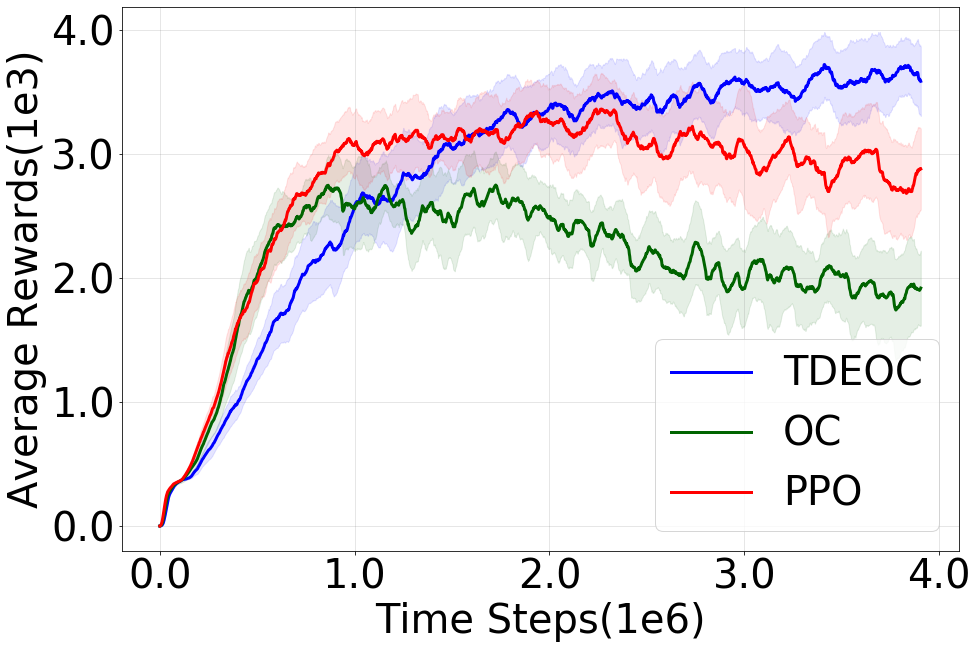} \label{Fig_Walker_results}} 
    \subfloat[TMaze(Discrete)]{\includegraphics[scale=0.12]{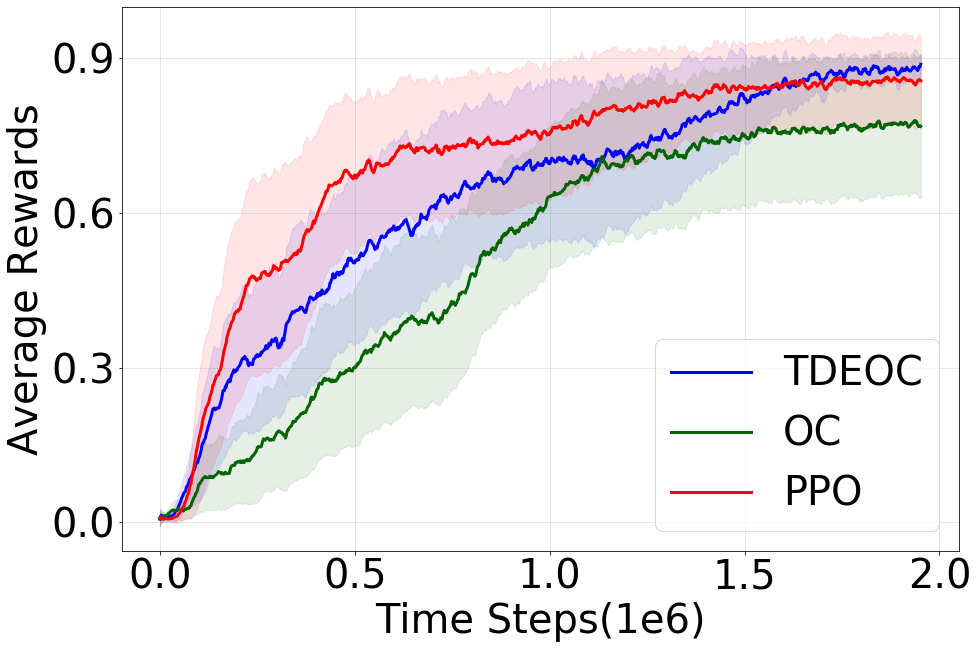}  \label{Fig_Tmaze_discrete_results}} \\
    % % \hrule
    \vspace{1mm}
    \textsc{\textbf{Option Relevance}}\\
    \subfloat[HalfCheetah-v2]{\includegraphics[scale=0.12]{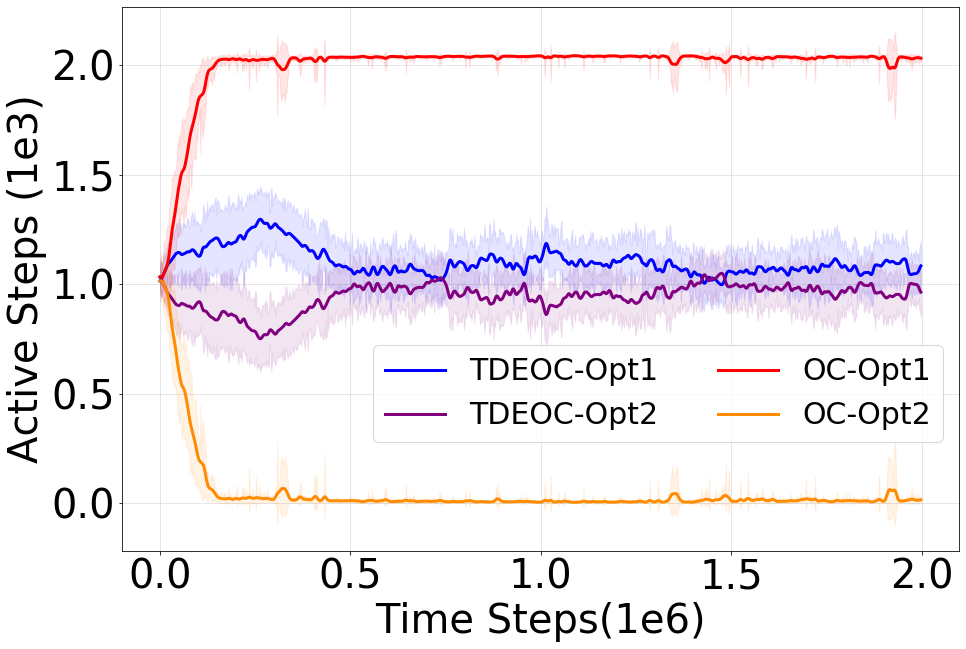}  \label{Fig_HalfCheetah_relevance}}
    \subfloat[Hopper-v2]{\includegraphics[scale=0.12]{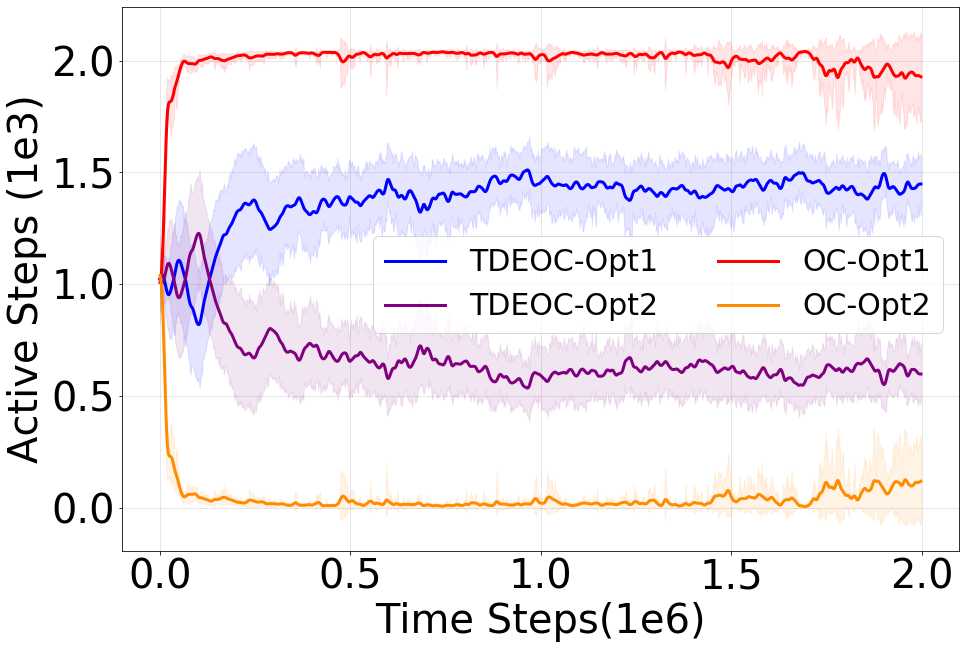} \label{Fig_Ant_relevance}} 
    \subfloat[Walker2d-v2]{\includegraphics[scale=0.12]{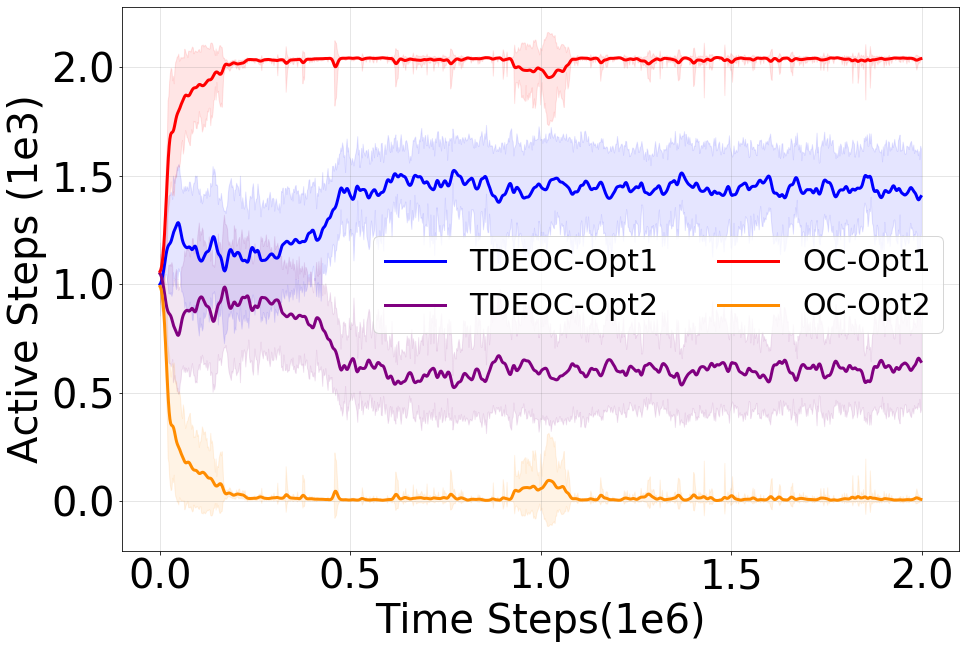} \label{Fig_Walker_relevance}} \\
    \caption{\textbf{TDEOC results on standard Mujoco and Miniworld tasks}. Our proposed termination objective significantly improves exploration, performance, and each option's relevance to the task. \textbf{Option activity refers to number of steps during which the option (Opt1 or Opt2) was active for buffer samples generated at respective time steps.} Each plot is averaged over 20 independent runs.
    }
    \label{Fig_TDEOC_results}
\end{figure*}
\setlength{\parindent}{0ex} \textbf{Tabular Four-rooms Navigation Task}\qquad
We first test our algorithm TDEOC, on the classic four-rooms navigation task \cite{Sutton:1999:MSF:319103.319108} where transfer capabilities of options were demonstrated against primitive action frameworks \cite{bacon2017option}. Initially the goal is located in the east hallway and the agent starts at a uniformly sampled state. After 1000 episodes, the goal state is moved to a random location in the lower right room. The goal state yields a reward of +1 while all other states produce no rewards.
% Implementation details are provided in Appendix \ref{APP_tabular}.
Visualizations of option terminations (Fig. \ref{Fig_Termination_plots}) show that 
% The termination probabilities of both Option-Critic (OC) and TDEOC are visualized in Fig \ref{Fig_Termination_Fourrooms_plots}. The darker colors represents higher termination probabilities while the darkest colors represent the walls of the environment. 
TDEOC identifies the hallways as the `\textit{bottleneck}' states where options tend to grow diverse. Fig. \ref{Fig_Termination_Fourrooms_plots} shows that both TDEOC and OC have nearly the same learning speed for the first 1000 episodes. Upon changing the goal state, TDEOC recovers faster than OC by almost 70 steps while exhibiting lower variance.\par
% \begin{figure*}[h]
%     \centering
%     \subfloat[HalfCheetah-v2]{\includegraphics[scale=0.13]{Figures/Option_activity_HalfCheetah.png}  \label{Fig_HalfCheetah_relevance}}
%     \subfloat[Ant-v2]{\includegraphics[scale=0.13]{Figures/Option_activity_Ant.png} \label{Fig_Ant_relevance}} 
%     \subfloat[Walker2d-v2]{\includegraphics[scale=0.13]{Figures/Option_activity_Walker.png} \label{Fig_Walker_relevance}} \\
%     \caption{
%     % \textbf{TDEOC results on standard Mujoco and Miniworld tasks}. Mujoco experiments are recorded until four million steps (1 iteration = 2048 steps), ten million steps for Humanoid-v2 and two million steps for Miniworld experiments as well as all option relevance plots. 
%     Option's activity refers to number of steps the option (Opt1 or Opt2) was active in an iteration. Each plot is averaged over 20 runs.}
%     \label{Fig_TDEOC_results}
% \end{figure*}
\begin{figure*}[!ht]
    \centering
     \subfloat[HalfCheetahHurdle-v0]{\includegraphics[scale=0.12]{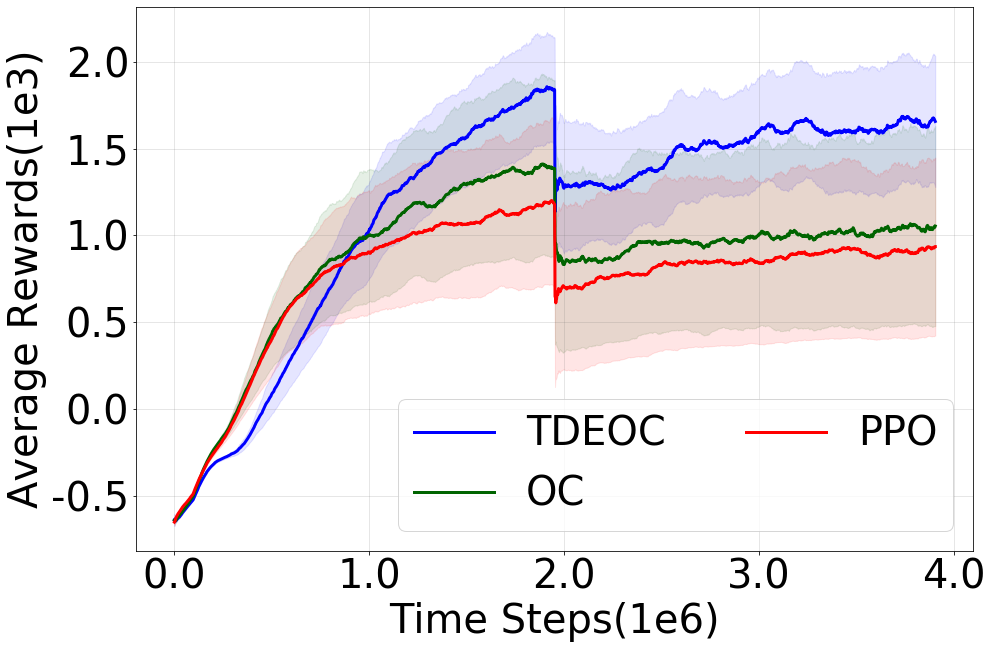}  \label{Fig_TDEOC_HalfCheetahHurdle}}
    \subfloat[HopperIceWall-v0]{\includegraphics[scale=0.12]{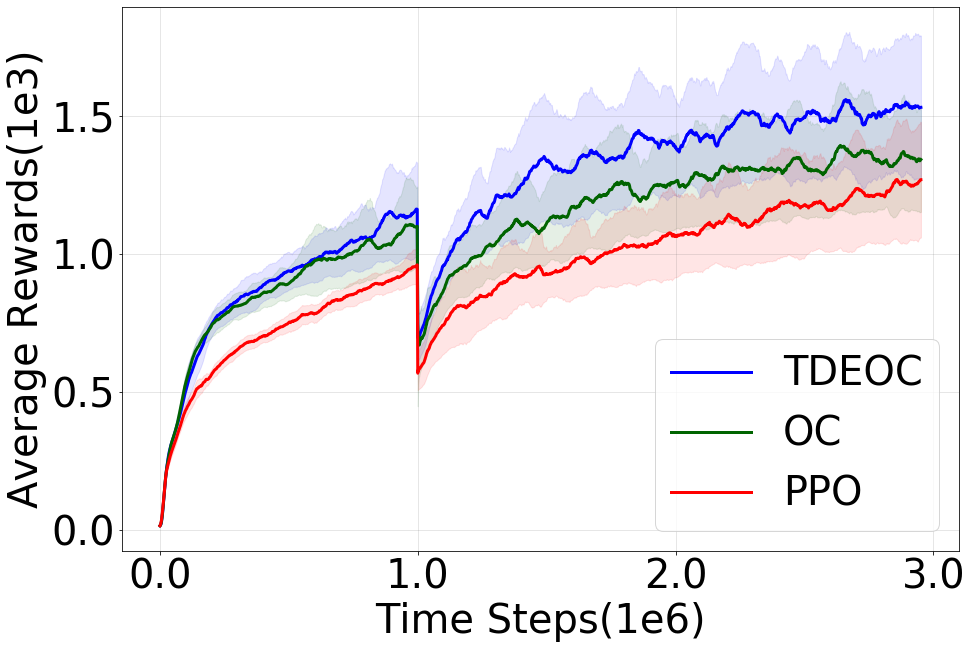} \label{Fig_TDEOC_HopperIce}} 
    \subfloat[TMaze (Continuous)]{\includegraphics[scale=0.12]{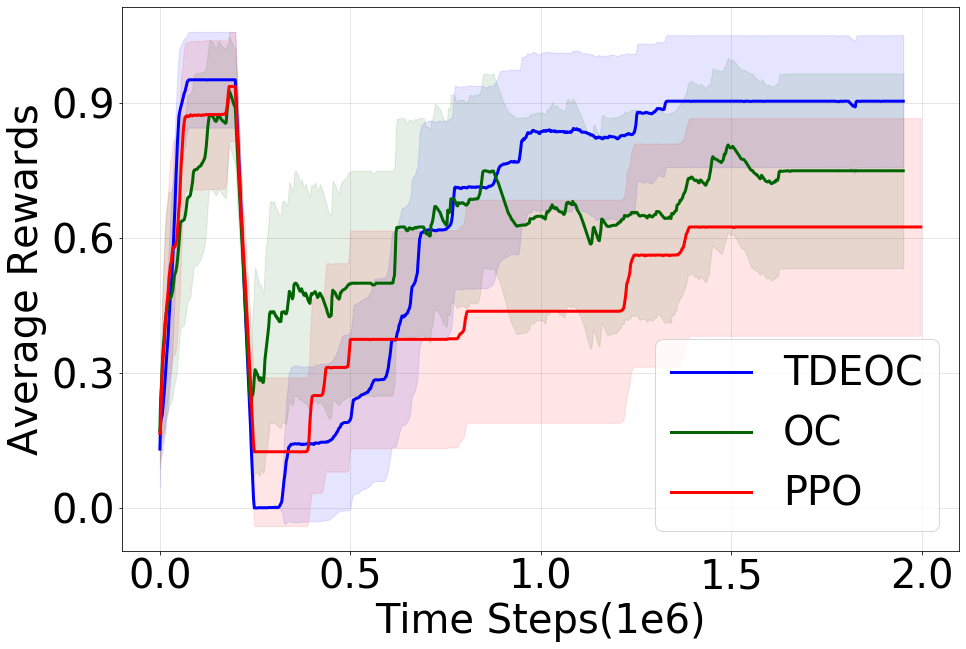}  \label{Fig_TDEOC_TMaze_transfer_results}} 
    \hspace{1mm} \\
     \subfloat[Steps$<$2e6]{\includegraphics[scale=0.225]{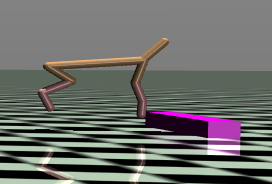}}\hspace{0.2mm}
    \subfloat[Steps$>$2e6]{\includegraphics[scale=0.275]{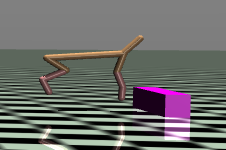}}\hspace{2.0mm}
    \subfloat[Steps$<$1e6]{\includegraphics[scale=0.084]{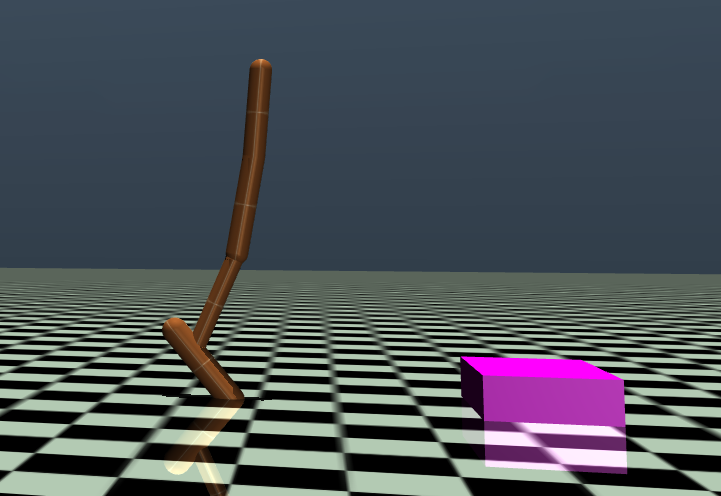}}\hspace{0.2mm}
    \subfloat[Steps$>$1e6]{\includegraphics[scale=0.0933]{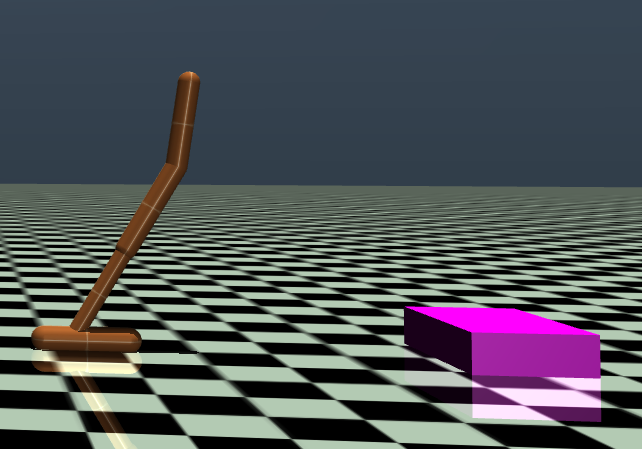}}\hspace{2.0mm}
    \subfloat[Steps$<$ 2e5]{\includegraphics[scale=0.0631]{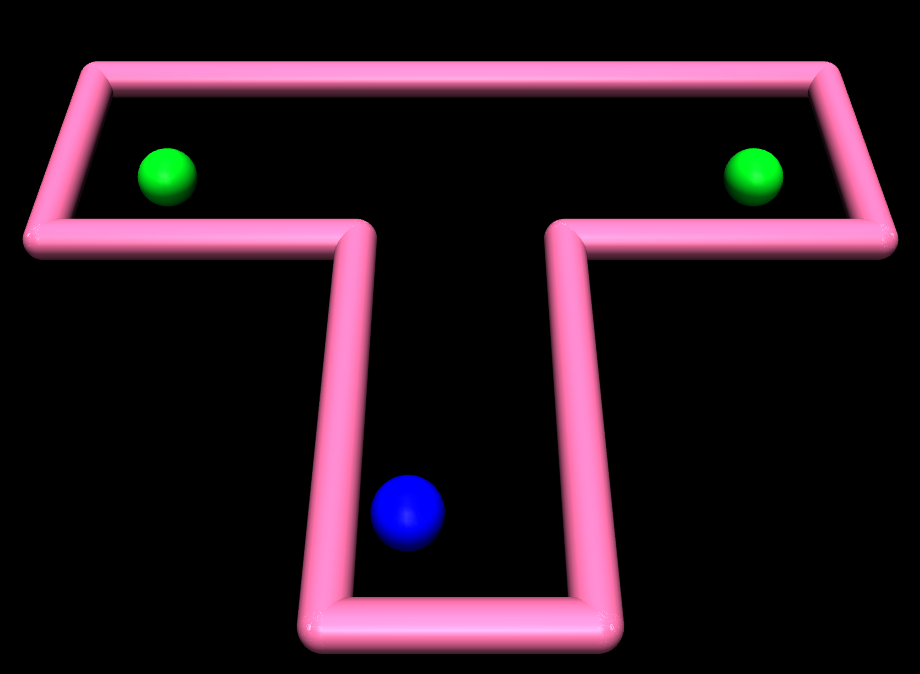}}\hspace{0.2mm}
    \subfloat[Steps$>$ 2e5]{\includegraphics[scale=0.0631]{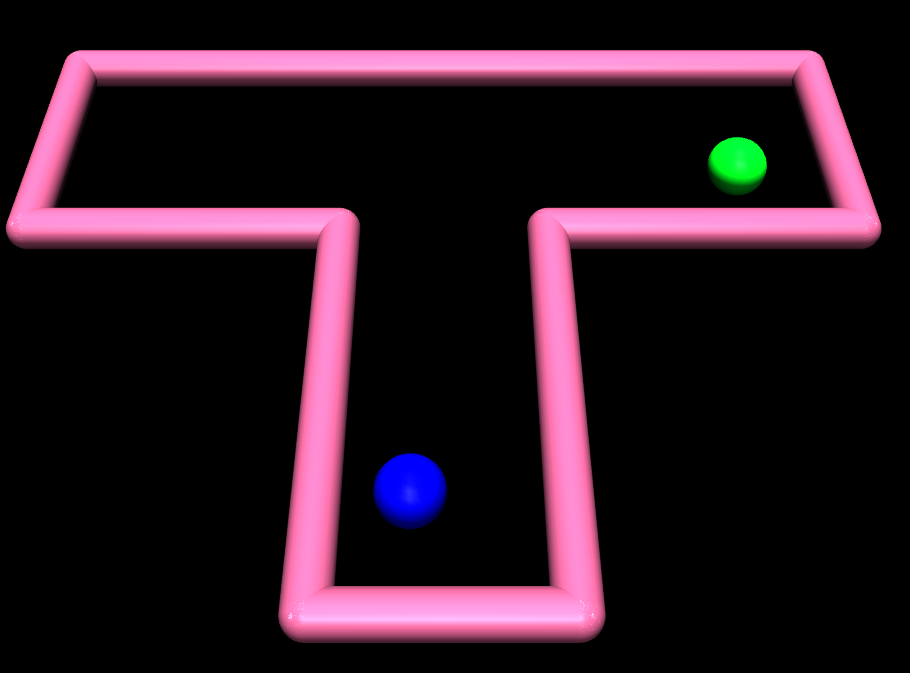}}
    \caption{\textbf{TDEOC results on three transfer tasks in Mujoco} each averaged over 20 independent runs. The height of the hurdle in HalfCheetahWall-v0 is increased by 0.8 metres after 2e6 steps. For HopperIceWall-v0, the block is moved 0.5 metres away from the agent's starting point after 1e6 steps. As for TMaze, the most frequent goal is removed after 2e5 steps. 
    % Results demonstrate TDEOC's ability to reuse specialized options to recover from changes in the environment.
    }
    \label{Fig_DEOC_term_transfer_mujojco_plots}
\end{figure*} 
\textbf{Continuous Control Tasks}\qquad
Next, we show the advantages of a diversity-targeted termination in the non-linear function approximation setting using standard Mujoco tasks \cite{todorov2012mujoco}. 
% Since PPOC uses a replay buffer to collect samples from the current policy we can easily standardize the $ \mathcal{R}_{bonus} $ samples. 
% We evaluate the algorithms on longer learning horizons (than in Fig \ref{Fig_DEOCvsPPOC}) to contrast stability. The average results from 20 runs are plotted (Fig \ref{Fig_TDEOC_results}). Details about the implementation and the hyper-parameters are provided in Appendix \ref{App_Nonlinearcase}. \\
We tested the performance of the TDEOC algorithm against Option-Critic (OC) and PPO \cite{Schulman2017ProximalPO}. 
% The results are presented in Fig. \ref{Fig_TDEOC_results}.  
Fig. \ref{Fig_TDEOC_results} shows that while OC quickly stagnates to a sub-optimal solution, TDEOC keep improving. We believe the reason for OC's stagnation is caused by sub-optimal option selection caused by terminations due to noisy value estimates. Since the sub-optimal option isn't adequately explored, it leads to the selection of a sub-optimal action, which can be catastrophic in states where \textit{balance} is vital.
% We believe the reason for OC's stagnation is sub-optimal option selection caused by terminating due to noisy value estimations. During the initial stages, one of the options quickly dominates  the entire task, preventing the other options to be explored. Consequently, any noise-induced option terminations occurring from then on would most likely lead to the selection of a sub-optimal option and a sub-optimal action which follows. Taking a \textit{bad} action can be catastrophic in states where `\textit{balance}' is vital. 
TDEOC, on the other hand, learns to generate diverse yet relevant option trajectories, thereby gaining better control. This explains why TDEOC handles environment perturbations more robustly. To demonstrate this property, we visualize the activity of each option for TDEOC and OC, in terms of the number of steps the option was active for buffer samples generated at respective time steps. (Fig. \ref{Fig_HalfCheetah_relevance}, \ref{Fig_Ant_relevance}, \ref{Fig_Walker_relevance}). Unlike OC, where only one option stays relevant for the task, TDEOC encourages both options to be selected fairly. TDEOC achieves a new state-of-the-art performance, not only outperforming OC by a wide margin, but also PPO. Our approach easily extends to very complex high dimensional tasks such as Humanoid-v2. TDEOC also exhibits lower variance demonstrating stable results across various random seeds despite using the same hyper-parameter settings.
See Appendix \ref{app_option_relevance} for additional results.
% Please refer to Appendix \ref{app_option_relevance} for further details and results on remaining tasks.
% PPOC uses entropy regularization for policy over option updates which encourages strong cooperative options.
TDEOC however, exhibits slower learning during the initial phase. This is to be expected, as TDEOC grooms both options to remain relevant and useful, while OC only updates a single dominant option, which requires fewer samples.
% TDEOC grooms both options to remain relevant without degeneration. Meanwhile, OC quickly prioritizes one of the options to perform the entire task, only requiring samples to update a single dominant option. 
% It is important to note that since diversity used for updates is relative to other states recorded in the buffer, the likelihood of options terminating at a state depends on how diverse their behavior is compared to other observed states. 
We study the \textit{critical states} which inspire diversity in Section \ref{section_Interpreting_options}.\\
\textbf{Sparse Reward Tasks}\qquad
We evaluate our approach in 3D visual control tasks implemented in Miniworld \cite{gym_miniworld} with discrete actions and a visual sensory input. We consider the T-Maze and Sidewalk environment. 
% Due to a sparse reward, as with the four-rooms task, we do not augment the reward with $ \mathcal{R}_{bonus}(S_{t}) $, but still compute it for termination updates. DEOC runs are also omitted due to the same. 
We observe that while OC stagnates to a sub-par solution for both tasks, TDEOC manages to learn a better solution faster (Fig. \ref{Fig_TDEOC_results}). TDEOC even outperforms PPO in Sidewalk, despite the added complexity of learning a hierarchy. We can also observe significantly lower variance in the TDEOC plots.
% The implementation details and hyper-parameters are provided in Appendix \ref{App_Nonlinearcase}. 
\subsection{Evaluating Transfer} \label{section_transfer_tasks}
A key advantage of using options is to learn skills which can be reused in similar tasks. In Section \ref{section_TDEOC_experiments}, we studied this property in the tabular four-rooms task. In this section, we further test our approach on tasks which require adapting to changes. The benefits of using options are best observed in tasks where hierarchical representation can be exploited. \\
\textbf{HalfCheetah Hurdle}\qquad 
Through this experiment we evaluate the ability of the agent to react to changes happening later in the trajectory. 
Reusing the HalfCheetah-v2 environment, a hurdle of height 0.12 metres is placed 10m away from the agent's starting position. After two million steps, the height of the hurdle is increased by 0.8 metres (to 2 metres). 
% To account for this change, the agent needs to adapt its leaping technique without flipping over before crossing the wall. 
Not only does TDEOC learn faster than option-critic and PPO (Fig. \ref{Fig_TDEOC_HalfCheetahHurdle}), it also adapts to the change quicker. TDEOC also keeps improving after recovery while PPO's and OC's performances stagnate. Despite OC being more robust to transfer \cite{bacon2017option}, PPO recovers faster than OC, as indicated by a smaller difference in performance after recovery than before. This suggests higher velocity can cause increased instability during recovery.  \\
% Such a stable adaptation is difficult to achieve especially when the agent has learned to perform the initial task with significant speed as achieved by TDEOC. \par
\textbf{Hopper Ice Wall}\qquad
We add a friction-less block in the Hopper-v2 task with dimensions (0.25m, 0.4m, 0.12m) corresponding to its length, width and height respectively, 2.8 metres away from the agent's starting position. 
% The agent needs to learn how to slide over this block without losing balance and resume hopping.
After a million steps, the block is moved 0.5 metres away from the agent's starting position. 
% Such a change requires the agent to re-evaluate its technique for jumping onto the block and not losing balance when making contact with it.
Initially,  TDEOC and option-critic have a similar rate of improvement. However, after the change, TDEOC learns to stabilize better and keeps improving (Fig. \ref{Fig_TDEOC_HopperIce}). \\
\textbf{TMaze Continuous}\qquad
% In Section \ref{section_continuous_control_single} we use the T-Maze environment with discrete action space and a visual input.
We use a task similar to the sparse reward task TMaze from Miniworld \cite{Khetarpal2020OptionsOI}. There are two goals located at both ends of the hallway, each producing a reward of +1. After 200,000 steps, the goal most visited is removed, forcing the agent to seek the other goal. Although OC initially recovers from the change better, TDEOC surpasses OC, achieving better final performance (Fig. \ref{Fig_TDEOC_TMaze_transfer_results}).
%We illustrate this qiual in Section \ref{section_Interpreting_options}.
\subsection{Interpreting Options Behavior} \label{section_Interpreting_options}
\begin{figure}[h]
    \centering
    \includegraphics[scale=0.20]{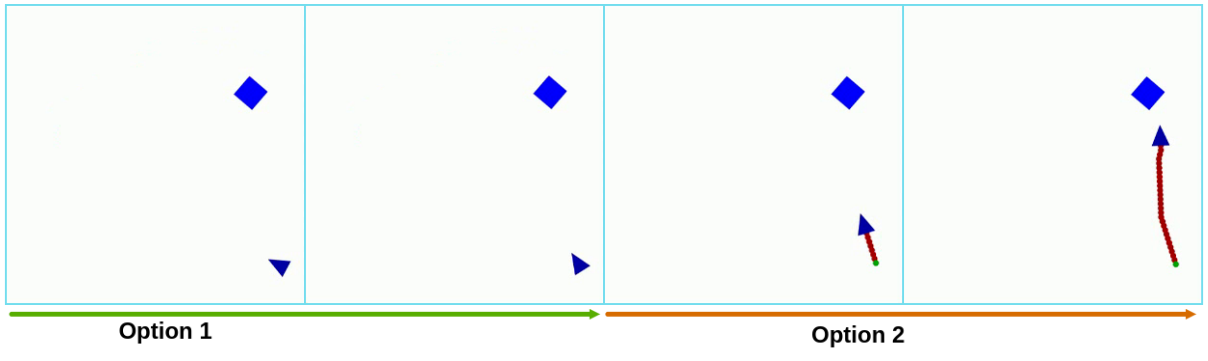}
    \caption{\textbf{Option trajectories in OneRoom task.} The first options scans the environment for the goal while the other option moves forward towards it.}
    \label{Fig_oneroom_trajectory}
\end{figure}

Learning tasks hierarchically through options can help us better understand the agent's solution. In this section, we study qualitatively the states targeted by TDEOC and the corresponding options behavior. Videos of all our experiments are provided on our website \footnote{\url{https://sites.google.com/view/deoc/home}}.\\
\begin{figure}
    \centering
    \subfloat[Trajectory before task change]{\includegraphics[scale=0.29]{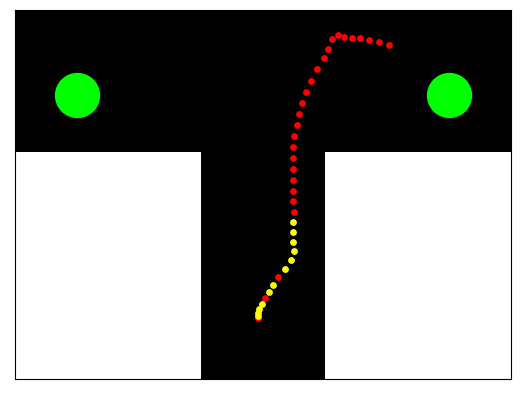}}\hspace{2mm}
    \subfloat[Trajectory after most frequent goal is removed]{\includegraphics[scale=0.21]{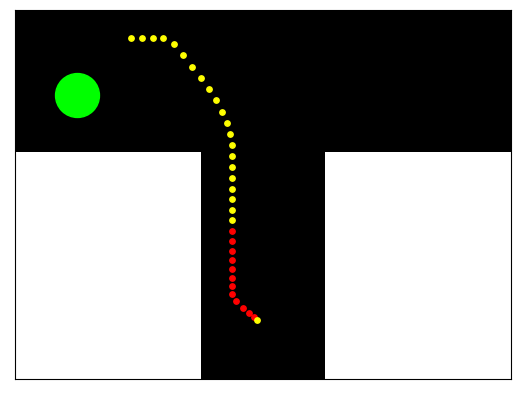}} \\
    \subfloat[Terminations for option 1 ($\beta_{o1}$)]{\includegraphics[scale=0.315]{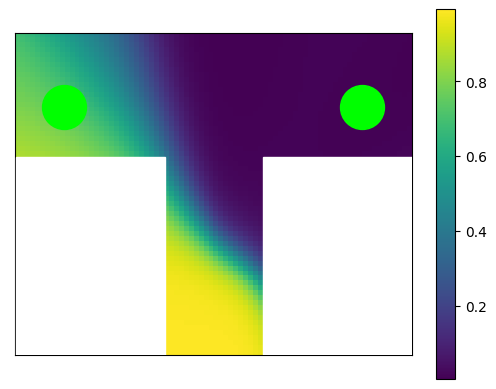}}\hspace{2mm}
    \subfloat[Terminations for option 2 ($\beta_{o2}$)]{\includegraphics[scale=0.315]{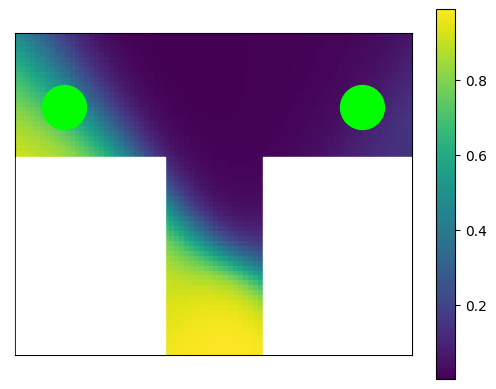}}\\
    % \subfloat[]{\includegraphics[scale=0.07]{Figures/TMaze_Both.png}}
    % \subfloat[]{\includegraphics[scale=0.07]{Figures/TMaze_Left.png}}
    \caption{\textbf{Visualizations on TMaze task using two options} (marked red and yellow respectively in (a) and (b)). Option terminations localize in the vertical hallway where the agent has yet to decide which goal to navigate towards.}
    \label{Fig_Tmaze_transfer_terminations}
\end{figure}
\begin{figure}[!h]
    \centering
    \includegraphics[scale=0.2]{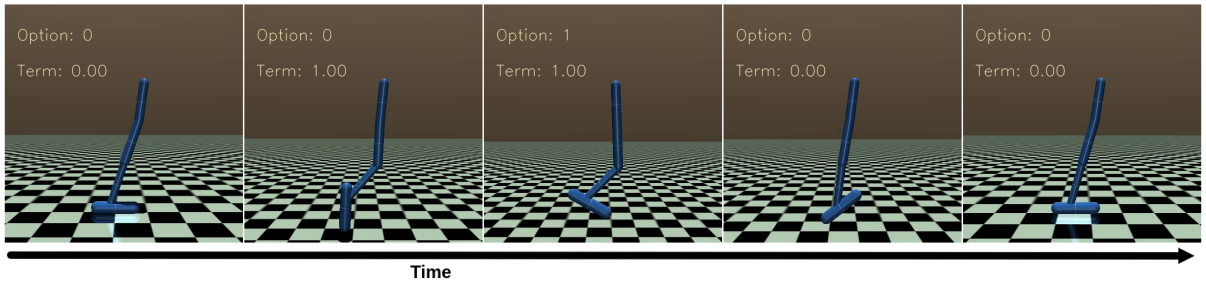}
    \caption{\textbf{Sample trajectory of the Hopper-v2 task.} Terminations are localized near states where the agent is in the air. Both options collaborate to ensure proper posture and balance prior to descending.}
    \label{Fig_hopper_trajectory}
\end{figure} 
\textbf{TMaze}\qquad
We visualize option behaviors and terminations for the TMaze transfer task from Section \ref{section_transfer_tasks}. Figure \ref{Fig_Tmaze_transfer_terminations} visualize a sample run where options terminate in the vertical hallway. Once the target goal is determined, a single option navigates towards it. This seems very intuitive, as the choice of navigating to either of the goals is still open in that hallway, indicating that the options are capable of diverse strategies each focused on a specific goal. Therefore, the termination objective we proposed  gives rise to intuitive and reusable option behaviors. \\
\textbf{Hopper-v2}\qquad
% In Section \ref{section_TDEOC_experiments}, we show that not only does TDEOC demonstrate better performance, it also handles perturbations in the environment better. We evaluate how two options collectively help stabilize the agent and observe the states where such stability is crucial. 
Next, we consider the Hopper-v2 simulation from Mujoco. From Fig. \ref{Fig_hopper_trajectory}, we can see that options terminate when the agent is in the air, just before descending. Naturally, landing without losing balance is very important, as even a slight mistake can cause catastrophic outcomes. TDEOC manages to employ both options around these states to complement each other, thereby achieving robust control and proper balance (See Appendix \ref{App_hopper}). \\
\textbf{OneRoom}\qquad
We consider the OneRoom task from Miniworld. We visualize a sample trajectory (Fig. \ref{Fig_oneroom_trajectory}) where one option learns to scan the room by turning on the spot, and upon observing the goal, the second option navigates towards it (see results in Appendix \ref{App_oneroom}). The termination objective hence gives rise to intuitive option strategies.
\section{Related work}
%In this paper, we propose a termination objective which attempts to promote options to remain relevant by exploiting diversity in options. 
Over the years, identifying bottleneck states as useful sub-goals for options has had lot of success \cite{McGovern_Automaticdiscovery, Stolle_learningoptions, Bacon2013OnTB}. Our approach can be seen to target bottleneck states characterized by diversity in the induced  option set. Unsupervised skill discovery using diversity has also shown to be capable of learning challenging tasks using an information theoretic objective \cite{gregor2016variational,diaynpaper-2018}. Our algorithm however, exploits a novel behavioral diversity metric for option discovery while also being capable of learning the task simultaneously. While most similar works use states to distinguish and specialize options \cite{gregor2016variational,diaynpaper-2018,termination-critic2019}, our approach exploits option's behavior.
%Our approach also generalizes to a wide range of tasks effortlessly.
Additionally, our approach can learn reusable and interpretable options even though we do not explicitly shape the initiation set~\cite{bagaria2020option, Khetarpal2020OptionsOI}.
% Intrinsic Motivation
Intrinsic motivation and reward modifications have been very successful in inducing certain desirable properties in RL algorithms, such as efficient exploration \cite{count_1, count_2, count_4, count_3}. Our approach falls in this category as well. A lot of literature relates to learning reward functions  to improve performance \cite{ng1999policy,zheng2018learning}, generalization and robustness \cite{singh2010intrinsically}, and we would like to investigate the utility of such methods in learning options as well.
\section{Discussions and Future Work}
In this paper, we highlighted the importance of learning a diverse set of options end-to-end. Inspired by intrinsic motivation, we presented Diversity-Enriched Option-Critic (DEOC),  and proposed a novel diversity-targeted, information theoretic termination objective capable of generating diverse option strategies to encourage exploration and make the learning more reliable. The new termination objective, coupled with DEOC's reward augmentation, produces relevant, robust and useful options. Our proposed algorithm, Termination-DEOC (TDEOC), significantly outperforms option-critic as well as PPO on a wide range of tasks. TDEOC can potentially benefit in scaling tasks to longer horizons. In the future, we would like to further investigate general methods for inferring reward functions and optimization criteria that lead to efficient exploration and transfer in new domains. 
% TDEOC manages to become one of the best performing model-free hierarchical reinforcement learning algorithm capable of robust learning and transfer while generating intuitive behavioral abstractions without any restrictions on option's initialization. \\ 
% A major restriction of our approach is that the sample complexity would naturally keep increasing as we increase the number of options for a task. An elegant solution to this would be to restrict the number of option choices at a given state using an attention-based approach such as interest functions. Such a strategy can easily help the algorithm learn longer horizon tasks with multiple events over the agent's trajectory. 
\nocite{puterman_book}
\nocite{bellman1954}
\nocite{henderson2017multitask}
\nocite{reproducibility_checklist}
% Do not set \bibliographystyle here: aaai21.sty already selects the required
% aaai21 style at \begin{document}. A second \bibliographystyle writes a
% conflicting \bibstyle to the .aux and makes BibTeX fail.
\bibliography{DEOC_paper}

@article{Sutton:1999:MSF:319103.319108,
  title={Between MDPs and semi-MDPs: A framework for temporal abstraction in reinforcement learning},
  author={Sutton, Richard S and Precup, Doina and Singh, Satinder},
  journal={Artificial intelligence},
  volume={112},
  number={1-2},
  pages={181--211},
  year={1999},
  publisher={Elsevier}
}

@inproceedings{parr1998reinforcement,
  title={Reinforcement learning with hierarchies of machines},
  author={Parr, Ronald and Russell, Stuart J},
  booktitle={Advances in neural information processing systems},
  pages={1043--1049},
  year={1998}
}

@article{dietterich2000hierarchical,
  title={Hierarchical reinforcement learning with the MAXQ value function decomposition},
  author={Dietterich, Thomas G},
  journal={Journal of artificial intelligence research},
  volume={13},
  pages={227--303},
  year={2000}
}

@article{mcgovern2001automatic,
  title={Automatic discovery of subgoals in reinforcement learning using diverse density},
  author={McGovern, Amy and Barto, Andrew G},
  booktitle={ICML 2001, volume 1, 361–368},
  year={2001}
}

@inproceedings{Precup2000TemporalAI,
  title={Temporal abstraction in reinforcement learning},
  author={Doina Precup and Richard S. Sutton},
  booktitle={ICML},
  year={2000}
}

@inproceedings{bacon2017option,
  title={The option-critic architecture},
  author={Bacon, Pierre-Luc and Harb, Jean and Precup, Doina},
  booktitle={Thirty-First AAAI Conference on Artificial Intelligence},
  year={2017}
}

@article{diaynpaper-2018,
  author    = {Benjamin Eysenbach and
               Abhishek Gupta and
               Julian Ibarz and
               Sergey Levine},
  title     = {Diversity is All You Need: Learning Skills without a Reward Function},
  journal   = {CoRR},
  volume    = {abs/1802.06070},
  year      = {2018},
  url       = {http://arxiv.org/abs/1802.06070},
  archivePrefix = {arXiv},
  eprint    = {1802.06070},
  bibsource = {dblp computer science bibliography, https://dblp.org}
}

@inproceedings{deliiberationcost,
  title={When waiting is not an option: Learning options with a deliberation cost},
  author={Harb, Jean and Bacon, Pierre-Luc and Klissarov, Martin and Precup, Doina},
  booktitle={Thirty-Second AAAI Conference on Artificial Intelligence},
  year={2018}
}

@InProceedings{TRIO,
  title = 	 {Time-Regularized Interrupting Options (TRIO)},
  author = 	 {Timothy Mann and Daniel Mankowitz and Shie Mannor},
  booktitle = 	 {Proceedings of the 31st International Conference on Machine Learning},
  pages = 	 {1350--1358},
  year = 	 {2014},
  editor = 	 {Eric P. Xing and Tony Jebara},
  volume = 	 {32},
  number =       {2},
  series = 	 {Proceedings of Machine Learning Research},
  address = 	 {Bejing, China},
  month = 	 {22--24 Jun},
  publisher = 	 {PMLR},
  url = 	 {http://proceedings.mlr.press/v32/mannb14.html}
}

@InProceedings{Stolle_learningoptions,
author="Stolle, Martin
and Precup, Doina",
editor="Koenig, Sven
and Holte, Robert C.",
title="Learning Options in Reinforcement Learning",
booktitle="Abstraction, Reformulation, and Approximation",
year="2002",
publisher="Springer Berlin Heidelberg",
address="Berlin, Heidelberg",
pages="212--223",
isbn="978-3-540-45622-3"
}

@article{haarnoja2018soft,
  title={Soft actor-critic: Off-policy maximum entropy deep reinforcement learning with a stochastic actor},
  author={Haarnoja, Tuomas and Zhou, Aurick and Abbeel, Pieter and Levine, Sergey},
  journal={arXiv preprint arXiv:1801.01290},
  year={2018}
}

@phdthesis{Bacon2013OnTB,
  title={On the bottleneck concept for options discovery},
  author={Bacon, Pierre-Luc},
  year={2013},
  school={Masters thesis, McGill University}
}

@phdthesis{Bacon2013phdthesis,
  title={Temporal Representation Learning},
  author={Bacon, Pierre-Luc},
  year={2018},
  school={PhD thesis, McGill University}
}

@article{Williams1991FunctionOU,
  title={Function optimization using connectionist reinforcement learning algorithms},
  author={Williams, Ronald J and Peng, Jing},
  journal={Connection Science},
  volume={3},
  number={3},
  pages={241--268},
  year={1991},
  publisher={Taylor \& Francis}
}

@inproceedings{Mnih2016AsynchronousMF,
  title={Asynchronous Methods for Deep Reinforcement Learning},
  author={Volodymyr Mnih and Adri{\`a} Puigdom{\`e}nech Badia and Mehdi Mirza and Alex Graves and Timothy P. Lillicrap and Tim Harley and David Silver and Koray Kavukcuoglu},
  booktitle={ICML},
  year={2016}
}

@article{Klissarov2017LearningsOE,
  title={Learnings Options End-to-End for Continuous Action Tasks},
  author={Martin Klissarov and Pierre-Luc Bacon and Jean Harb and Doina Precup},
  journal={ArXiv},
  year={2017},
  volume={abs/1712.00004}
}

@article{termination-critic2019,
  title={The Termination Critic},
  author={Anna Harutyunyan and Will Dabney and Diana Borsa and Nicolas Manfred Otto Heess and R{\'e}mi Munos and Doina Precup},
  booktitle={AISTATS},
  year={2019}
}

@inproceedings{McGovern_Automaticdiscovery,
 author = {McGovern, Amy and Barto, Andrew G.},
 title = {Automatic Discovery of Subgoals in Reinforcement Learning Using Diverse Density},
 year = {2001},
 isbn = {1558607781},
 publisher = {Morgan Kaufmann Publishers Inc.},
 address = {San Francisco, CA, USA},
 booktitle = {Proceedings of the Eighteenth International Conference on Machine Learning},
 pages = {361–368},
 numpages = {8},
 series = {ICML ’01}
}

@inproceedings{
bagaria2020option,
title={Option Discovery using Deep Skill Chaining},
author={Akhil Bagaria and George Konidaris},
booktitle={International Conference on Learning Representations},
year={2020},
url={https://openreview.net/forum?id=B1gqipNYwH}
}

@inproceedings{Vezhnevets2017FeUdalNF,
  title={FeUdal Networks for Hierarchical Reinforcement Learning},
  author={Alexander Sasha Vezhnevets and Simon Osindero and Tom Schaul and Nicolas Manfred Otto Heess and Max Jaderberg and David Silver and Koray Kavukcuoglu},
  booktitle={ICML},
  year={2017}
}

@article{Khetarpal2020OptionsOI,
  title={Options of Interest: Temporal Abstraction with Interest Functions},
  author={Khimya Khetarpal and Martin Klissarov and Maxime Chevalier-Boisvert and Pierre-Luc Bacon and Doina Precup},
  journal={Proceedings of the Thirty-Fourth AAAI Conference on Artificial Intelligence (AAAI-20)},
  year={2020},
}

@inproceedings{ng1999policy,
  title={Policy invariance under reward transformations: Theory and application to reward shaping},
  author={Ng, Andrew Y and Harada, Daishi and Russell, Stuart},
  booktitle={ICML},
  volume={99},
  pages={278--287},
  year={1999}
}

@inproceedings{count_1,
  title={Unifying count-based exploration and intrinsic motivation},
  author={Bellemare, Marc and Srinivasan, Sriram and Ostrovski, Georg and Schaul, Tom and Saxton, David and Munos, Remi},
  booktitle={Advances in Neural Information Processing Systems},
  pages={1471--1479},
  year={2016}
}

@inproceedings{count_2,
  title={Count-based exploration with neural density models},
  author={Ostrovski, Georg and Bellemare, Marc G and van den Oord, A{\"a}ron and Munos, R{\'e}mi},
  booktitle={Proceedings of the 34th International Conference on Machine Learning-Volume 70},
  pages={2721--2730},
  year={2017},
  organization={JMLR. org}
}

@article{count_3,
  title={Count-based exploration with the successor representation},
  author={Machado, Marlos C and Bellemare, Marc G and Bowling, Michael},
  journal={Proceedings of the 34th AAAI Conference on Artificial Intelligence (AAAI 2020)},
  year={2018}
}

@article{count_4,
  title={OpenAI Xi Chen, Yan Duan, John Schulman, Filip DeTurck, and Pieter Abbeel.\# exploration: A study of count-based exploration for deep reinforcement learning},
  author={Tang, Haoran and Houthooft, Rein and Foote, Davis and Stooke, Adam},
  journal={Advances in neural information processing systems},
  volume={30},
  pages={2753--2762},
  year={2017}
}

@inproceedings{zheng2018learning,
  title={On learning intrinsic rewards for policy gradient methods},
  author={Zheng, Zeyu and Oh, Junhyuk and Singh, Satinder},
  booktitle={Advances in Neural Information Processing Systems},
  pages={4644--4654},
  year={2018}
}

@article{singh2010intrinsically,
  title={Intrinsically motivated reinforcement learning: An evolutionary perspective},
  author={Singh, Satinder and Lewis, Richard L and Barto, Andrew G and Sorg, Jonathan},
  journal={IEEE Transactions on Autonomous Mental Development},
  volume={2},
  number={2},
  pages={70--82},
  year={2010},
  publisher={IEEE}
}

@inproceedings{todorov2012mujoco,
  title={Mujoco: A physics engine for model-based control},
  author={Todorov, Emanuel and Erez, Tom and Tassa, Yuval},
  booktitle={2012 IEEE/RSJ International Conference on Intelligent Robots and Systems},
  pages={5026--5033},
  year={2012},
  organization={IEEE}
}

@book{sutton2018reinforcementbook,
  title={Reinforcement learning: An introduction},
  author={Sutton, Richard S and Barto, Andrew G},
  year={2018},
  publisher={MIT press}
}

@article{bellman1954,
author = "Bellman, Richard",
fjournal = "Bulletin of the American Mathematical Society",
journal = "Bull. Amer. Math. Soc.",
month = "11",
number = "6",
pages = "503--515",
publisher = "American Mathematical Society",
title = "The theory of dynamic programming",
url = "https://projecteuclid.org:443/euclid.bams/1183519147",
volume = "60",
year = "1954"
}

@book{puterman_book,
author = {Puterman, Martin L.},
title = {Markov Decision Processes: Discrete Stochastic Dynamic Programming},
year = {1994},
isbn = {0471619779},
publisher = {John Wiley \& Sons, Inc.},
address = {USA},
edition = {1st}
}

@misc{baselines,
  author = {Dhariwal, Prafulla and Hesse, Christopher and Klimov, Oleg and Nichol, Alex and Plappert, Matthias and Radford, Alec and Schulman, John and Sidor, Szymon and Wu, Yuhuai and Zhokhov, Peter},
  title = {OpenAI Baselines},
  year = {2017},
  publisher = {GitHub},
  journal = {GitHub repository},
  howpublished = {\url{https://github.com/openai/baselines}},
}

@article{Schulman2017ProximalPO,
  title={Proximal Policy Optimization Algorithms},
  author={John Schulman and Filip Wolski and Prafulla Dhariwal and Alec Radford and Oleg Klimov},
  journal={ArXiv},
  year={2017},
  volume={abs/1707.06347}
}

@misc{gym_miniworld,
  author = {Chevalier-Boisvert, Maxime},
  title = {gym-miniworld environment for OpenAI Gym},
  year = {2018},
  publisher = {GitHub},
  journal = {GitHub repository},
  howpublished = {\url{https://github.com/maximecb/gym-miniworld}},
}

@article{gregor2016variational,
  title={Variational intrinsic control},
  author={Gregor, Karol and Rezende, Danilo Jimenez and Wierstra, Daan},
  journal={arXiv preprint arXiv:1611.07507},
  year={2016}
}

@article{henderson2017multitask,
   author = {{Henderson}, P. and {Chang}, W.-D. and {Shkurti}, F. and {Hansen}, J. and 
	{Meger}, D. and {Dudek}, G.},
    title = {Benchmark Environments for Multitask Learning in Continuous Domains},
  journal = {ICML Lifelong Learning: A Reinforcement Learning Approach Workshop},
     year={2017}
}

@misc{reproducibility_checklist,
    author = {Pineau, Joelle},
    title= {Machine Learning Reproducibility Checklist},
    year={2019},
    howpublished={\url{https://www.cs.mcgill.ca/~jpineau/ReproducibilityChecklist.pdf}},

}


\onecolumn
\appendix

\section{Proof Theorem \ref{terminationtheorem}}\label{app_proofterminationtheorem} 
\textit{Given 
% the gradient-based option-critic algorithm \cite{bacon2017option}, 
a set of Markov options $\Omega$ each with a stochastic termination function defined by Eq. \eqref{eq_deocobjective} and stochastic intra-option policies, with $|\Omega|<\infty$ and $|\mathcal{A}|<\infty$, repeated application of policy-options evaluation and improvement \cite{bacon2017option, Bacon2013phdthesis} yields convergence to a locally optimum solution. }\\

We continue the same assumptions and proofs derived by \citet{bacon2017option, Bacon2013phdthesis}. The termination functions ($\beta$) and the intra-option policies ($\pi_{o}$) are differentiable in their respective parameters, $\nu$ and $\theta$. Also, the parameters of the termination functions, $\nu$, and that of the intra-option policies, $\theta$, are independent \citep{bacon2017option}. We use the TDEOC algorithm with tabular intra-option Q-learning as a template for our proof. An $\epsilon$-greedy policy over options $\pi_{\Omega}$ is used. We reuse the notations and terminologies defined by \citet{bacon2017option}. The option-value function is defined as:
\begin{equation}
    Q_{\Omega}(s,o) = \sum_{a \in \mathcal{A}} \pi_{o}(a|s) Q_{U}(s,o,a)
\end{equation}
$Q_{U}: S \times \Omega \times \mathcal{A} \rightarrow \mathbb{R}$  is the value of executing the option in the context of state-option pair. $Q_{U}$ is evaluated using the one-step off-policy update defined as:
\begin{equation}\label{eq_q_update}
    Q_{U}(s,o,a) \leftarrow Q_{U}(s,o,a) + \alpha \Big[ r(s,a) + \gamma \big[ (1 -  \beta_{o}(s'))Q_{\Omega}(s',o) + \beta_{o}(s') \max_{\overline{o}} Q_{\Omega}(s',\overline{o}) \big] \Big]
\end{equation}
where $\alpha$ is the step size and the update target, $\delta$ is given by:
\begin{equation*}
\delta = r(s,a) + \gamma \big[ (1 -  \beta_{o}(s'))Q_{\Omega}(s',o) + \beta_{o}(s') \max_{\overline{o}} Q_{\Omega}(s',\overline{o}) \big] 
\end{equation*}
Now we present the intra-option policy optimization equations for option improvement. The objective to be maximized is the expected discounted return:
\begin{equation}
    \mathcal{J}(\theta) = Q_{\Omega}(s,o)
\end{equation}
where $\theta$ represents the parameters of the intra-option policies. \citet{bacon2017option} have derived the intra-option gradient \citep{bacon2017option,Bacon2013phdthesis} for option policy improvement as:
\begin{equation}
    \frac{\partial \mathcal{J}(\theta)}{\partial \theta} = \sum_{s.o} \mu_{\Omega}(s,o|s_{0},o_{0}) \sum_{a} \frac{\partial log \pi_{o}(a | s)}{\partial \theta} Q_{U}(s, o, a)
\end{equation}
where $\mu_{\Omega}(s_{0}, o_{0})$ is the discounted weighting of state-option pair along trajectories starting from $(s_{0}, o_{0})$. Under assumption:
\begin{equation}\label{eq_step_size}
\sum_{t=0}^{\infty} \alpha^{\{\theta\}}_{t} = \infty \:\: and \:\: \sum_{t=0}^{\infty} \big[\alpha^{\{\theta\}}_{t}\big]^{2} < \infty
\end{equation}
then the sequence $\{\theta_{t}\}_{t=0}^{\infty}$
\begin{equation}
    \theta_{t+1} \leftarrow \theta_{t} + \alpha_{t}^{\{\theta\}} \frac{\partial log \pi_{o_{t}}(a_{t} | s_{t})}{\partial \theta} Q_{U}(s_{t}, o_{t}, a_{t})
        % \theta_{\beta} \leftarrow \theta_{\beta} +  \alpha_{\theta_{\beta}} \frac{\partial \beta_{o_{t}}(s_{t+1})}{\partial \nu} \mathcal{D}(s_{t+1})
\end{equation}
converges with probability 1 and the $\lim_{t \rightarrow \infty} \frac{\partial \mathcal{J}(\theta)}{\partial \theta} = 0$ \citep{bacon2017option, Bacon2013phdthesis}.
As for the one-step off-policy intra-option Q learning update in Eq. \eqref{eq_q_update}, we can apply the same convergence properties used by \citet{sutton2018reinforcementbook} to show $Q_{U}$ converges to optimal behavior, $Q^{*}_{U}$ with probability 1 under the same conditions on sequence of step size as in Eq. \eqref{eq_step_size}. To understand it intuitively, the intra-option gradient updates converges intra-option policies to a locally optimum behavior while the $\epsilon$-greedy policy over options, $\pi_{\Omega}$,  greedily selects the option with the highest value upon termination.

Hence, our termination condition still allows policy-option iteration to converge to a locally optimum solution. The rate of convergence, however, may vary depending on the termination function.
The termination function during option evaluation is used to evaluate the value function $Q_{U}(s_{t}, o_{t}, a_{t})$, based on the stochastic likelihood of the current option continuing or terminating at the given state. Upon termination, the option with the highest value at that state is selected for evaluation. The structure of the termination function is irrelevant for evaluating and improving options, and is solely used to define the scope of option's abstraction. \citet{Bacon2013phdthesis} also derived a proof showing that a fixed termination function leave the convergence properties unaffected. The advantages of our proposed termination objective are mainly empirical. The assumptions $|\Omega|<\infty$ and $|\mathcal{A}|<\infty$ are required to ensure the intrinsic reward and the termination function are bounded.

\section{Effects of Varying the Number of Options}\label{app_variable_options}
\begin{figure}[!h]
    \centering
    \subfloat[Ant-v2]{\includegraphics[scale=0.11]{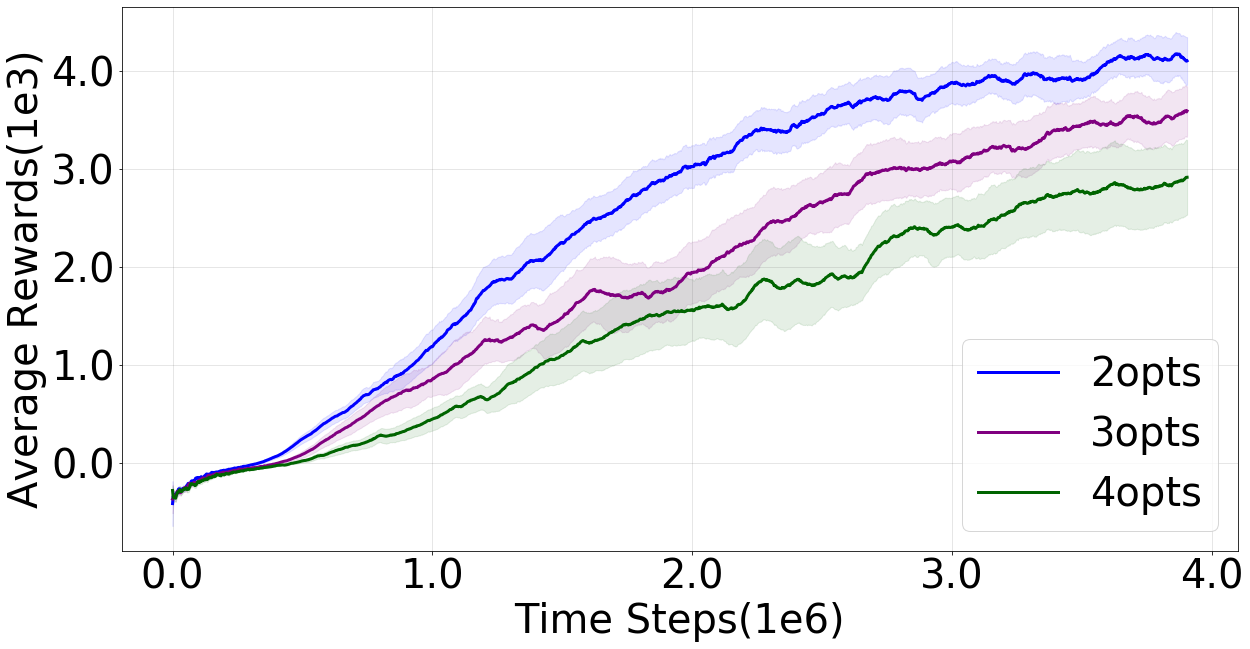}} 
    \subfloat[HalfCheetah-v2]{\includegraphics[scale=0.11]{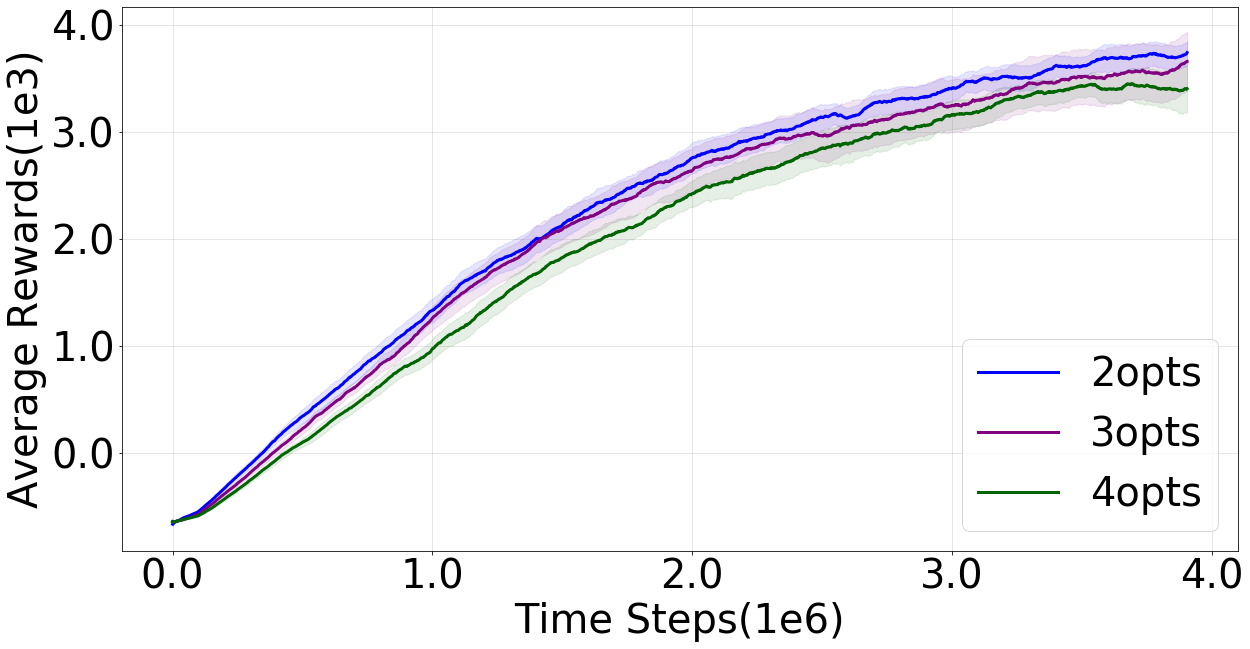}} \\
    \vspace{6mm}
    \subfloat[Hopper-v2]{\includegraphics[scale=0.11]{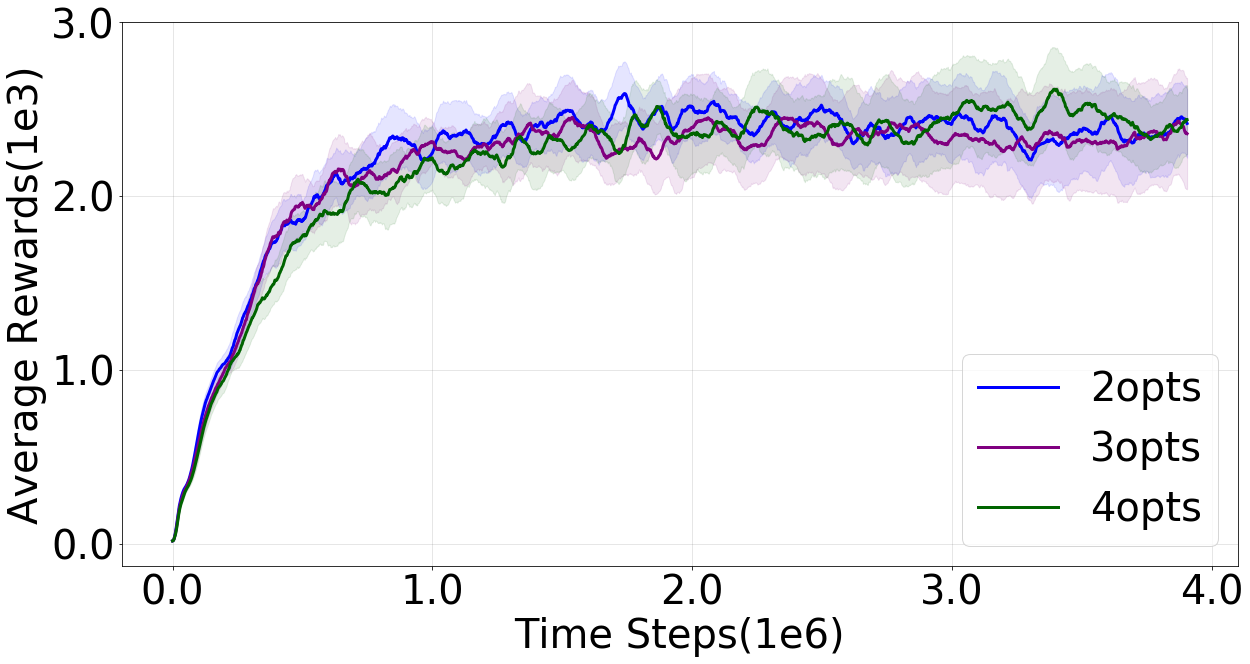}}
    \subfloat[Walker2d-v2]{\includegraphics[scale=0.11]{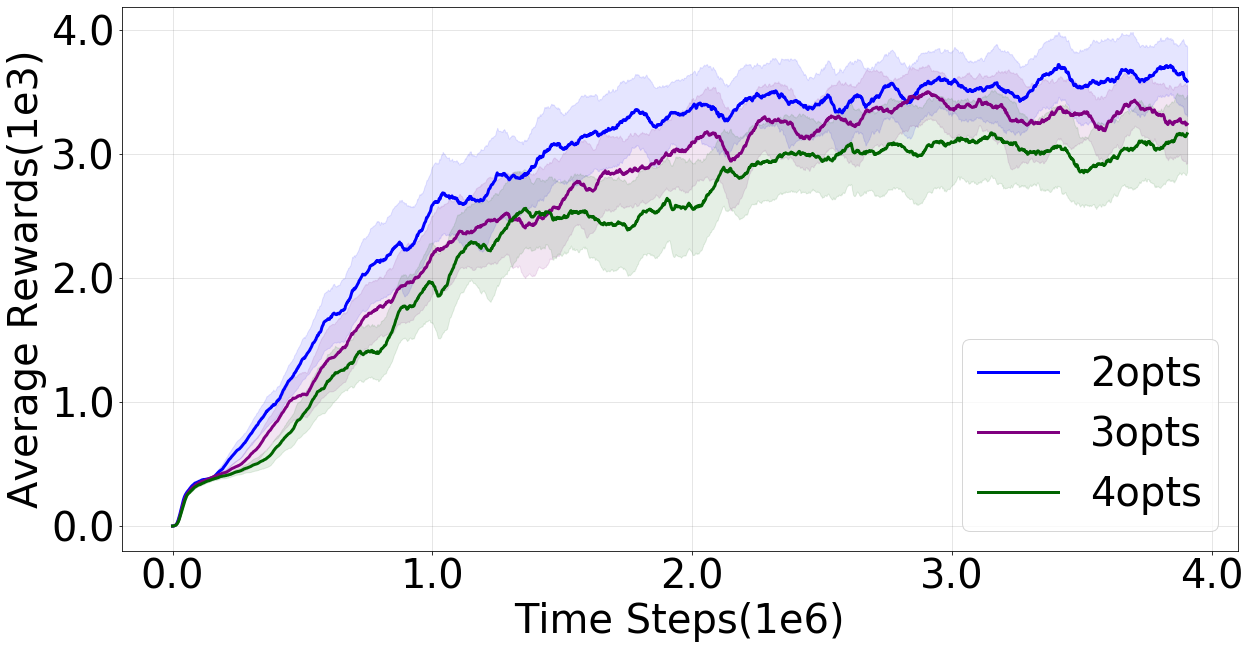}} 
    \caption{\textbf{TDEOC results on four Mujoco tasks with varying number of options.} Sample complexity keeps growing with increasing the number of options. Each line is an average of 20 runs.}
    \label{Fig_TDEOC_results_all_opts}
\end{figure}

\begin{figure}[!h]
    \centering
    \subfloat[Ant-v2]{\includegraphics[scale=0.11]{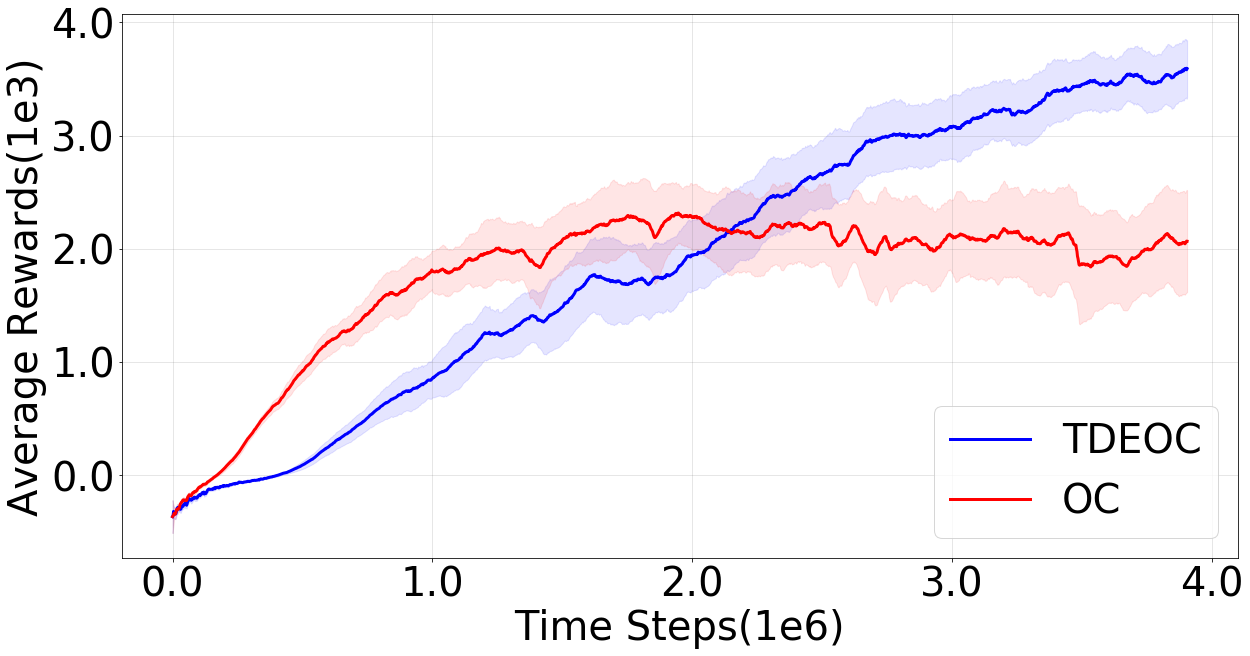} } 
    \subfloat[HalfCheetah-v2]{\includegraphics[scale=0.11]{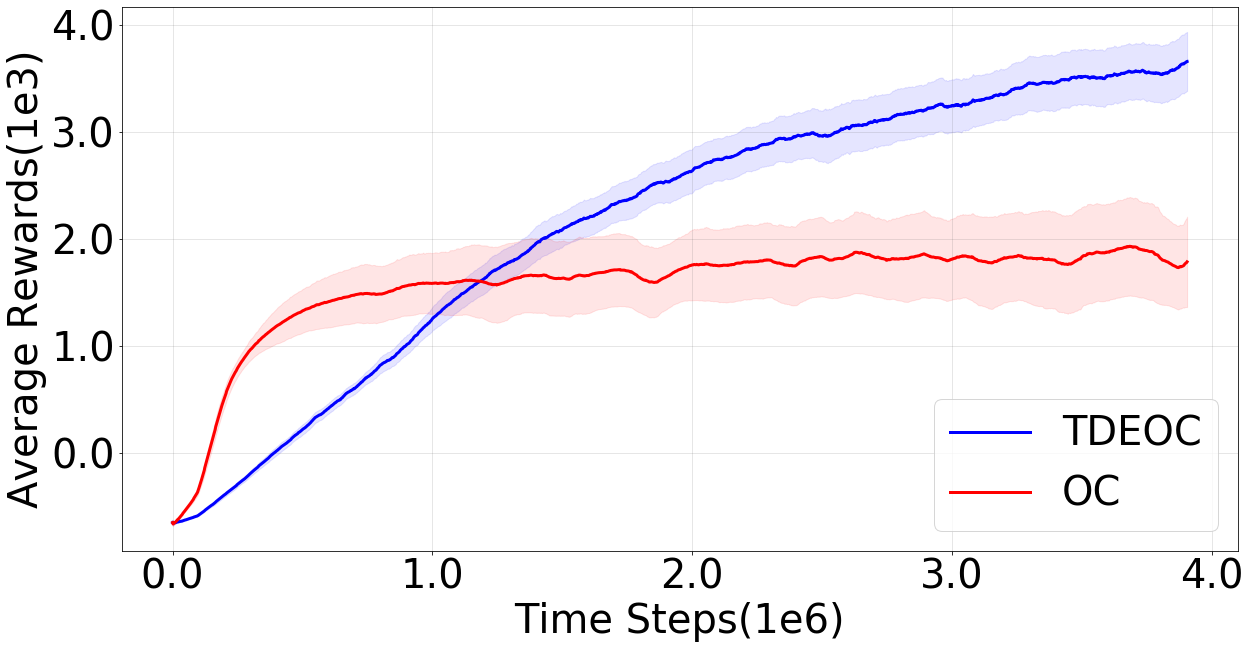} } \\
    \vspace{6mm}
    \subfloat[Hopper-v2]{\includegraphics[scale=0.11]{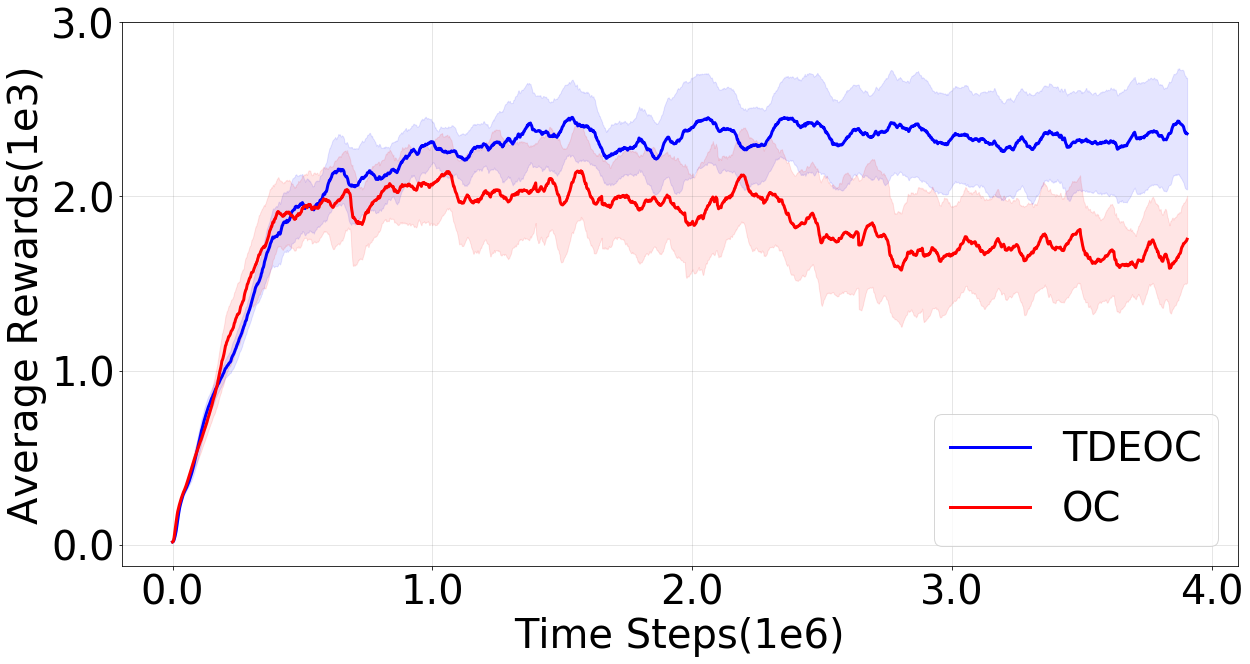} }
    \subfloat[Walker2d-v2]{\includegraphics[scale=0.11]{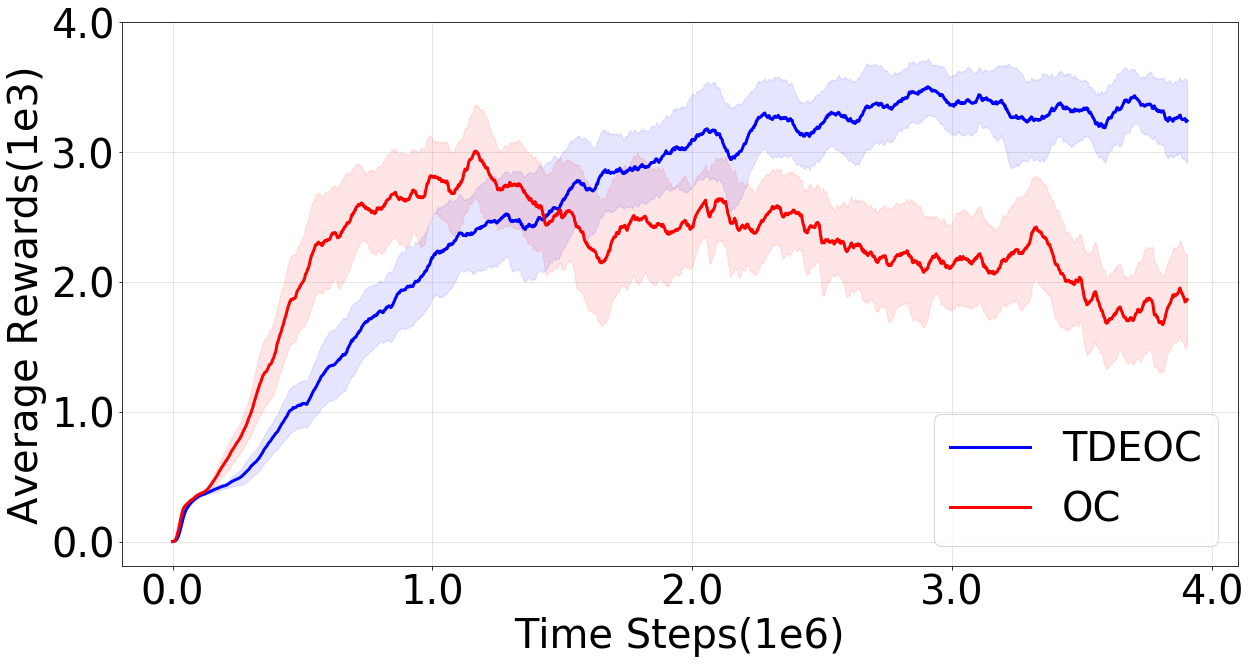}} 
    \caption{\textbf{TDEOC and OC results on four Mujoco tasks with three options.} Each line is an average of 20 runs.}
    \label{Fig_TDEOCvsOC_results_3opts}
\end{figure}

\begin{figure}[!h]
    \centering
    \subfloat[Ant-v2]{\includegraphics[scale=0.11]{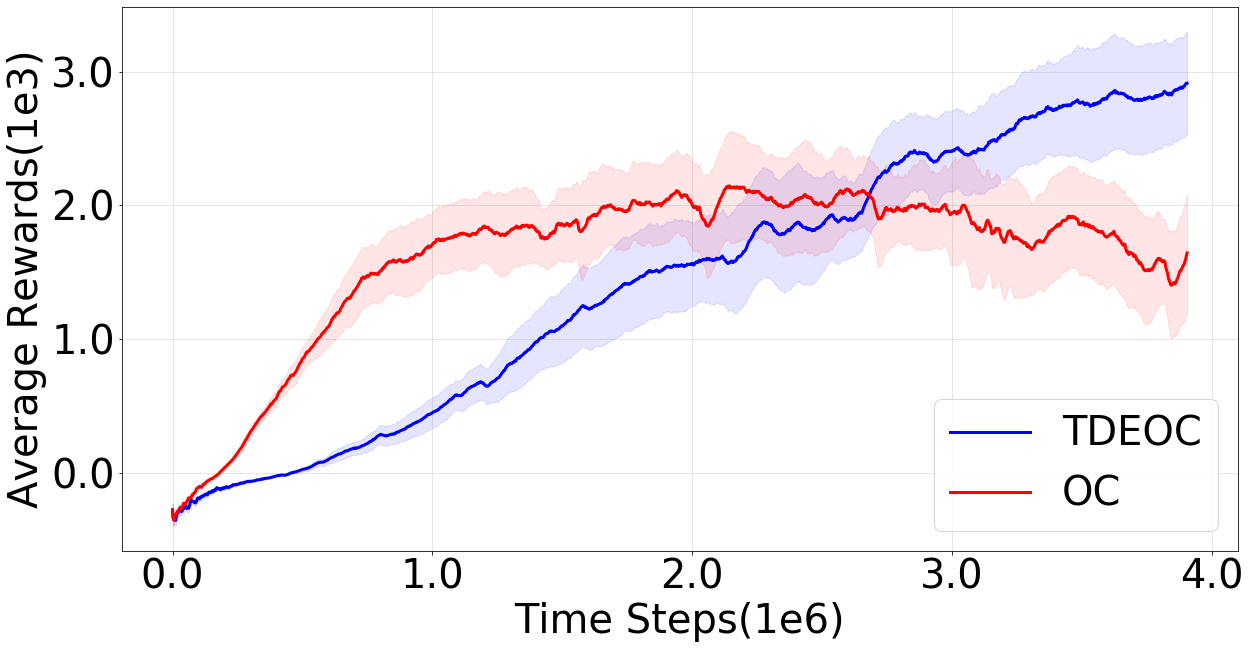} } 
    \subfloat[HalfCheetah-v2]{\includegraphics[scale=0.11]{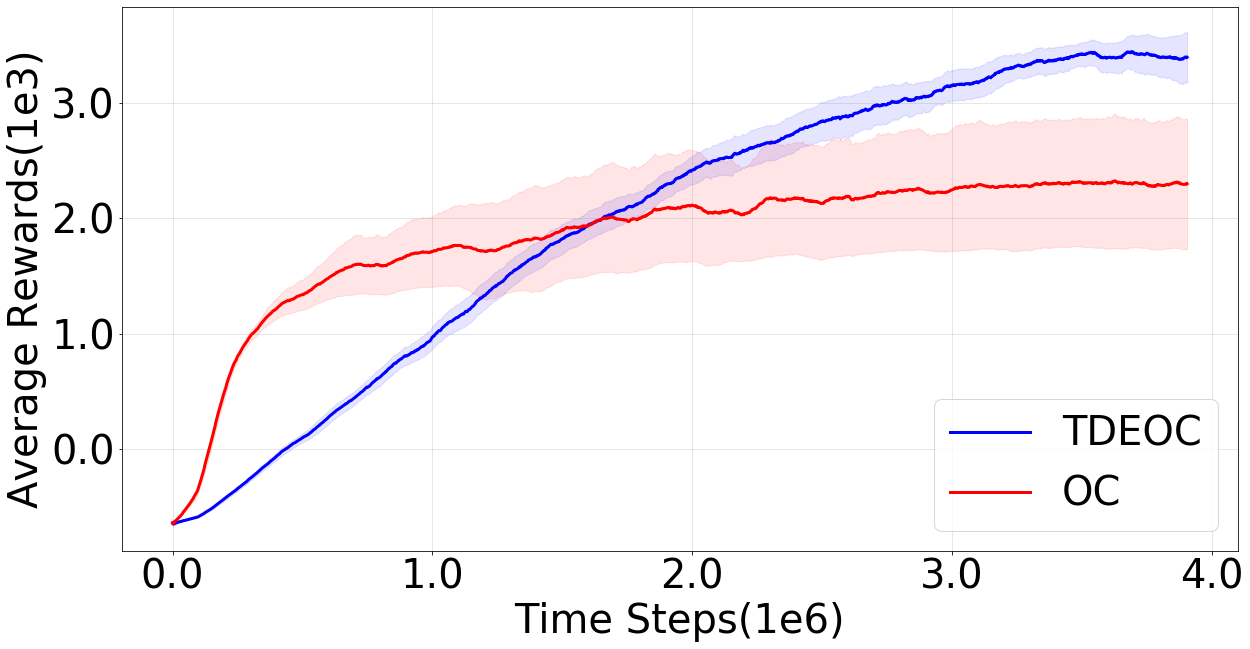} } \\
    \vspace{6mm}
    \subfloat[Hopper-v2]{\includegraphics[scale=0.11]{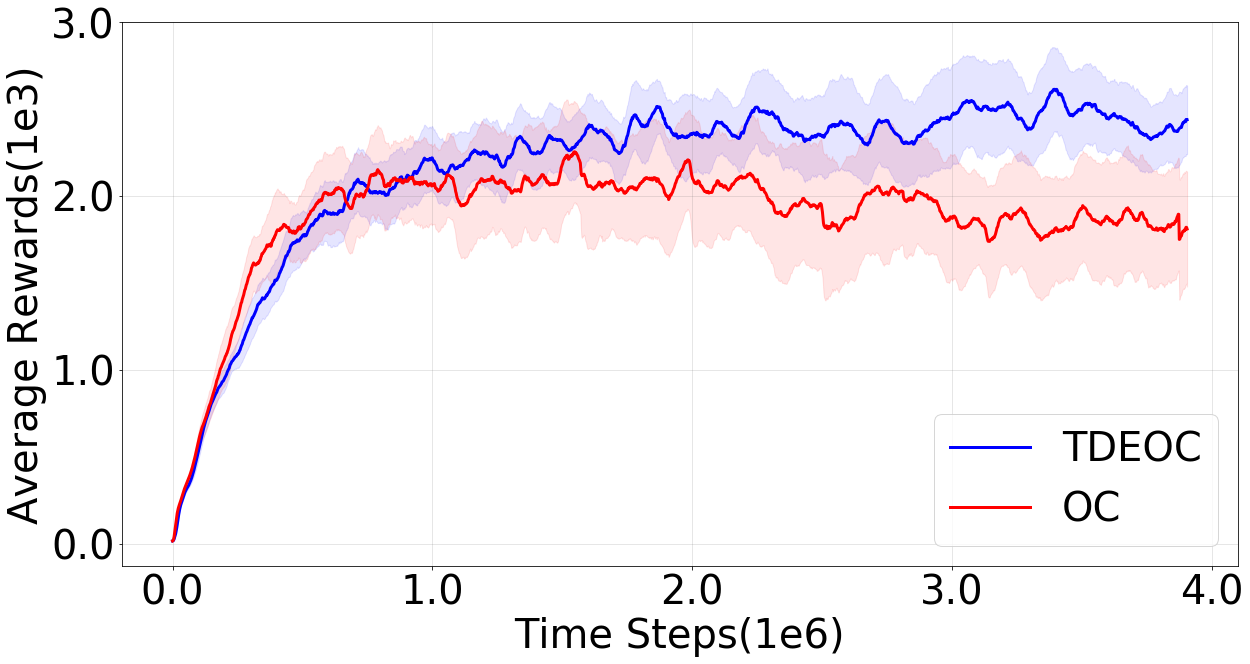}}
    \subfloat[Walker2d-v2]{\includegraphics[scale=0.11]{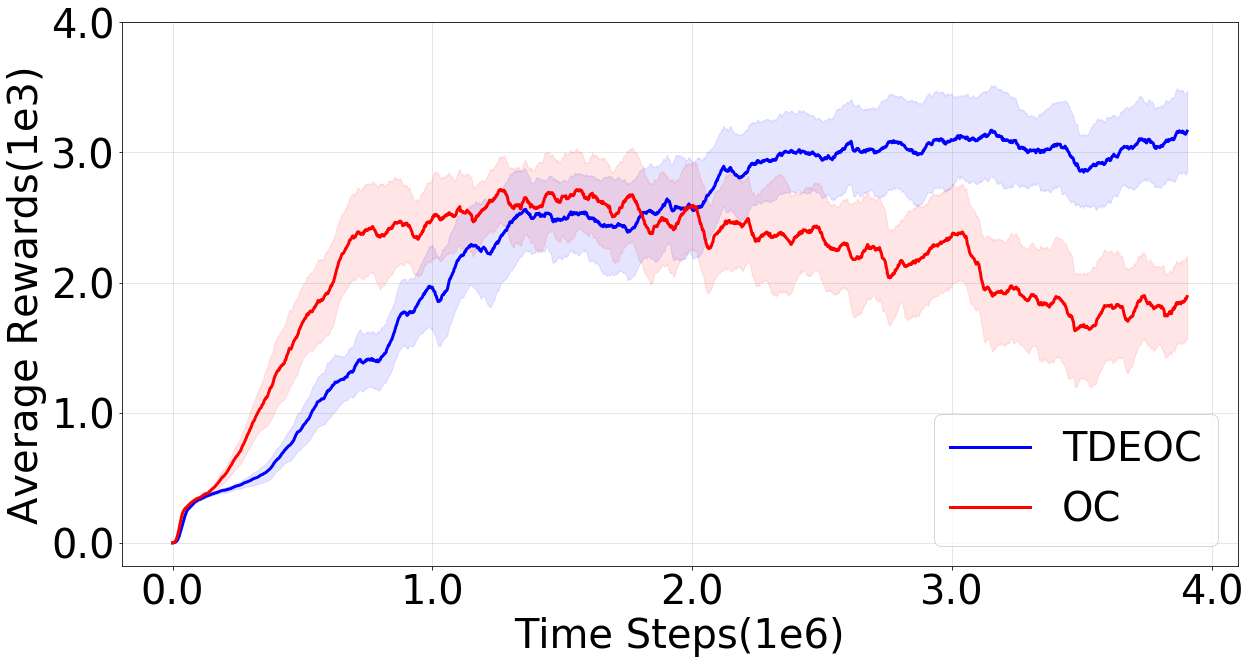}} 
    \caption{\textbf{TDEOC and OC results on four Mujoco tasks with four options.} Each line is an average of 20 runs.}
    \label{Fig_TDEOCvsOC_results_4opts}
\end{figure}

Option-critic framework assumes the number of options to be learnt, to be a hyper-parameter. To ensure reproducible results and fair comparisons, the choice of this hyper-parameter for all experiments presented in the paper relied mainly on previously tested experiments (refer Appendix \ref{App_Implementation_details}). We reuse the PPOC codebase \cite{baselines,Klissarov2017LearningsOE} for all our experiments involving non-linear function approximation. PPOC has only been tested on two options \cite{Klissarov2017LearningsOE} as of yet. However, to demonstrate that our algorithm TDEOC, easily generalizes to most tasks and implementation settings effortlessly, we study its performance while varying the number of options on standard Mujoco tasks. TDEOC's ability to encourage all available options to remain relevant to the task affects sample complexity. From Fig. \ref{Fig_TDEOC_results_all_opts} we can observe that sample complexity increases as the number of options increase. However, TDEOC still shows similar learning curves and properties despite varying the number of options and still outperforms OC using same number of options (Fig. \ref{Fig_TDEOCvsOC_results_3opts} and \ref{Fig_TDEOCvsOC_results_4opts}). 
\begin{figure}[!h]
    \centering
    \subfloat[Ant-v2]{\includegraphics[scale=0.11]{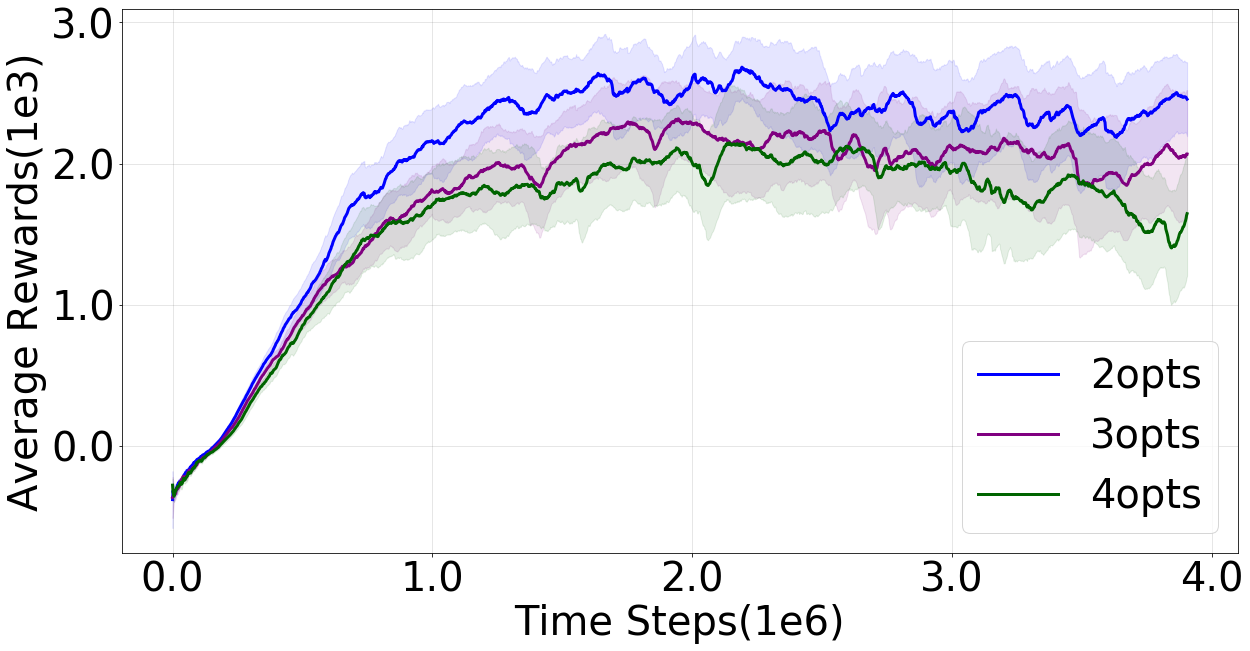} } 
    \subfloat[HalfCheetah-v2]{\includegraphics[scale=0.11]{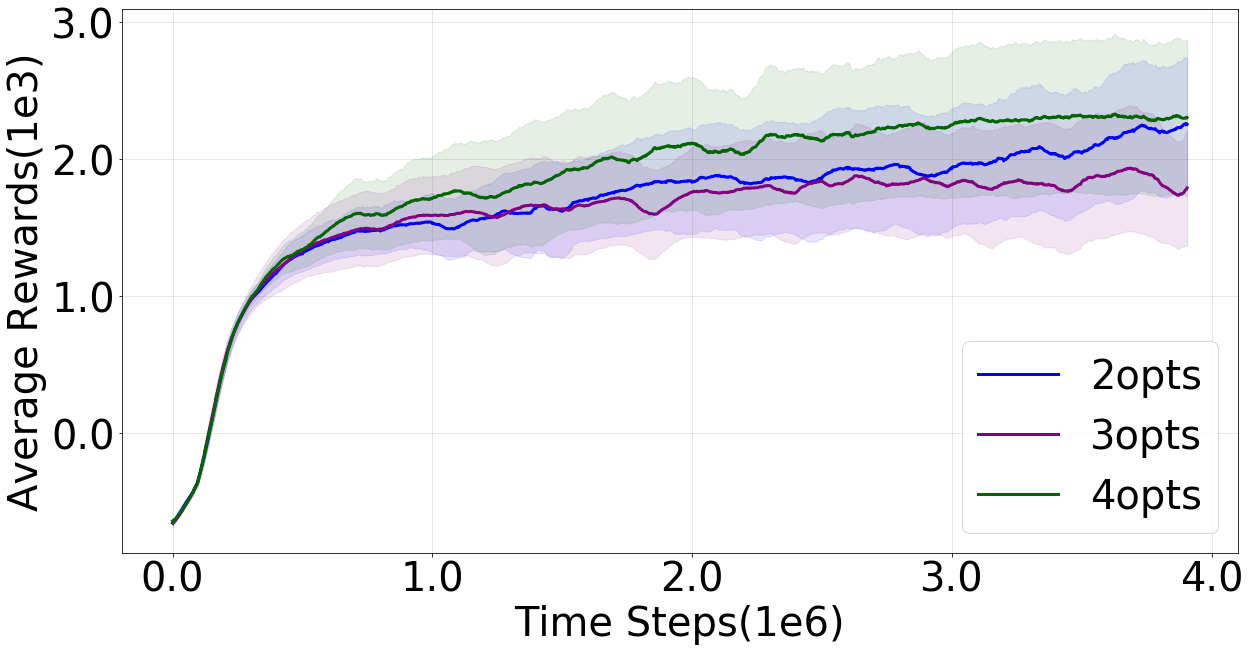} } \\
    \vspace{6mm}
    \subfloat[Hopper-v2]{\includegraphics[scale=0.11]{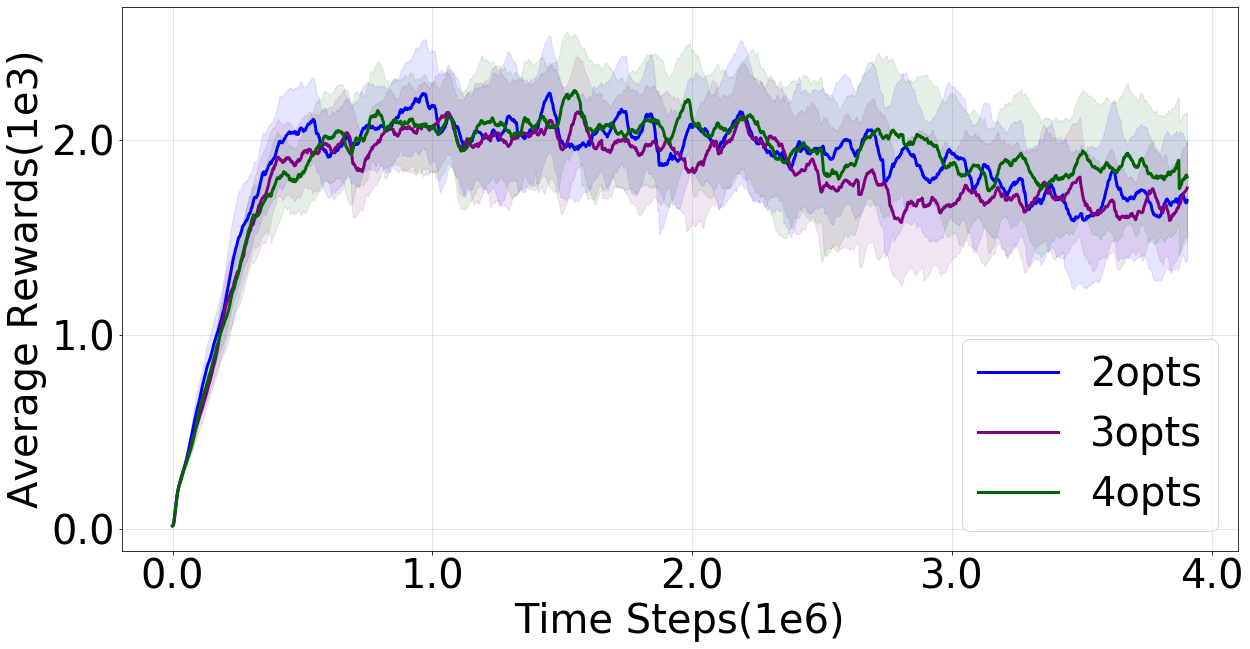} }
    \subfloat[Walker2d-v2]{\includegraphics[scale=0.11]{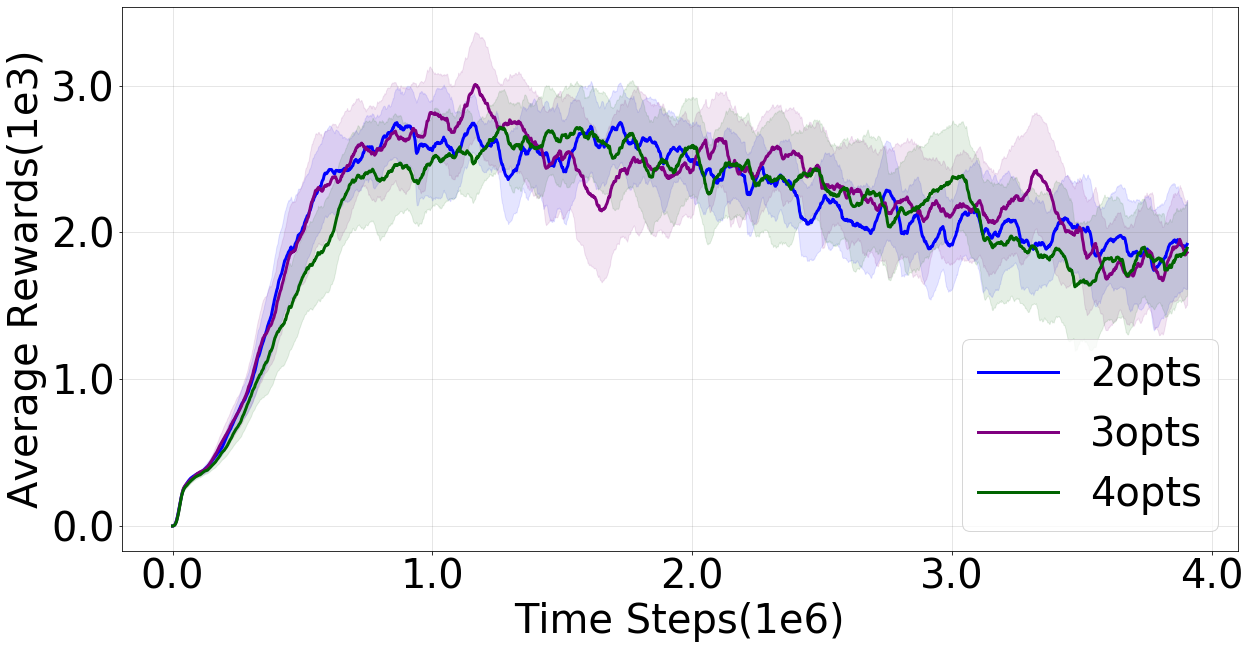} } 
    \caption{\textbf{Option-Critic results on four Mujoco tasks with varying number of options.} Each line is an average of 20 runs.}
    \label{Fig_OC_results_all_opts}
\end{figure}

\section{Option Relevance} \label{app_option_relevance}
In the paper we emphasize the importance of options simultaneously learning to remain relevant and useful for the task. In this section we empirically demonstrate how our approach of using a diversity motivated termination objective encourages both options to be explored fairly, thereby preventing a single option dominating over the entire task. 
% The biggest difference is observed in Mujoco tasks where one of the options quickly dominates over the entire task. 
In this section, we contrast our algorithm (TDEOC) using our proposed termination objective against Option-Critic(OC). We plot each option's activity in respective tasks for both algorithms. Option's activity corresponds to the number of steps it was active for buffer samples generated at respective time steps. Option 1 (indicated with \textit{Opt1} in the plots) refers to the dominant option in each task while option 2 (\textit{Opt2}) is the option which was active for fewer steps compared to option 1 for each run. Each line is an average of 20 runs. It is evident that TDEOC manages to explore and select both options much better than OC. The difference is significant in Mujoco tasks where one option quickly dominates over the entire task. We can reason TDEOC's performance in Fig. \ref{Fig_TDEOC_results} with how TDEOC exploits both the diverse options to to achieve better performance and stability.
\begin{figure*}[!h]
    \centering
    \subfloat[Humanoid-v2]{\includegraphics[scale=0.19]{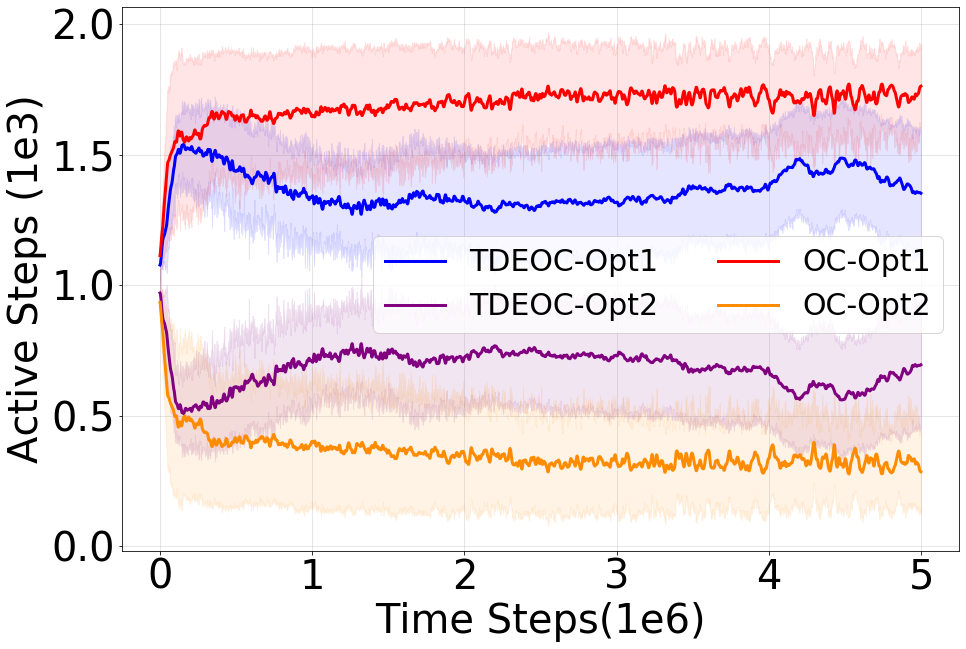} \label{Fig_Humanoid_results}} 
     \subfloat[Ant-v2]{\includegraphics[scale=0.19]{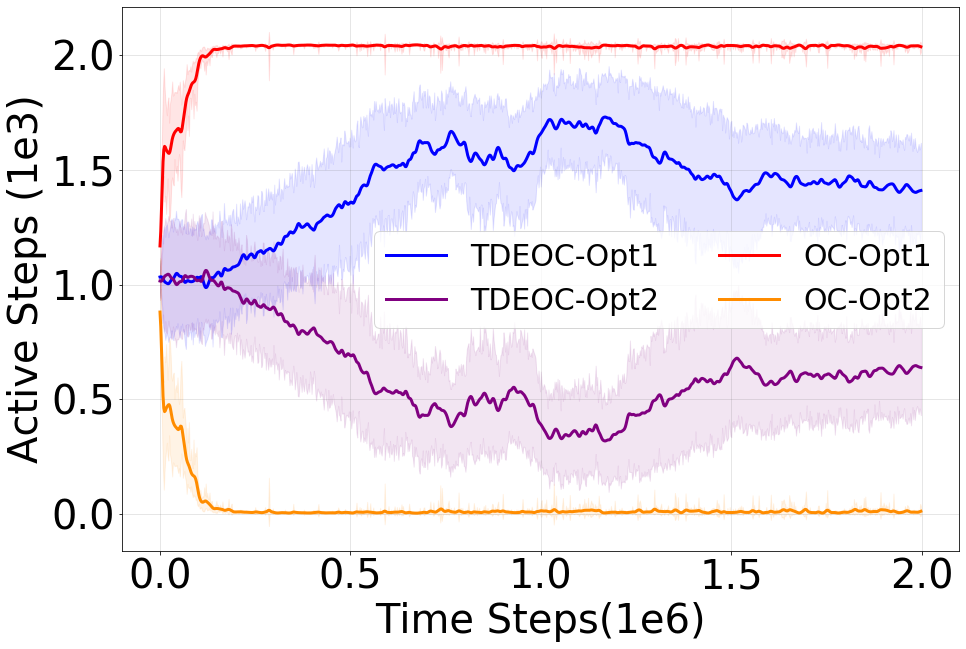} \label{Fig_hopper_results}}.\\
     \vspace{2mm}
    \subfloat[Sidewalk (Discrete)]{\includegraphics[scale=0.19]{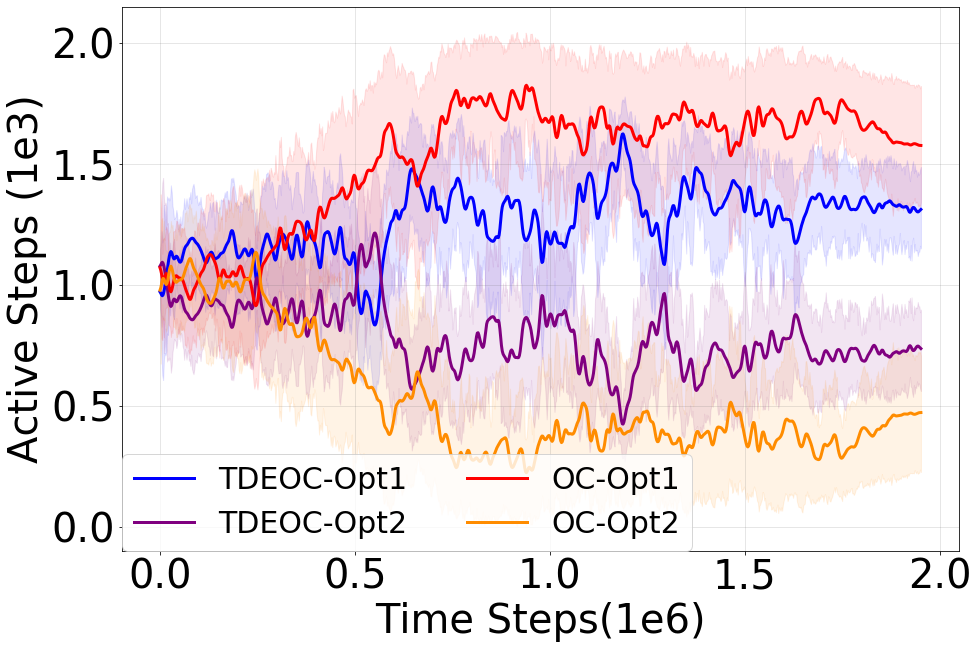}\label{Fig_Sidewalk_results}}\\
    \vspace{2mm}
    % \subfloat[Ant-v2]{\includegraphics[scale=0.13]{Figures/Option_activity_Ant.png} \label{Fig_Ant_results}} 
    % \subfloat[Walker2d-v2]{\includegraphics[scale=0.13]{Figures/Option_activity_Walker.png} \label{Fig_Walker_results}} 
    \subfloat[TMaze(Discrete)]{\includegraphics[scale=0.19]{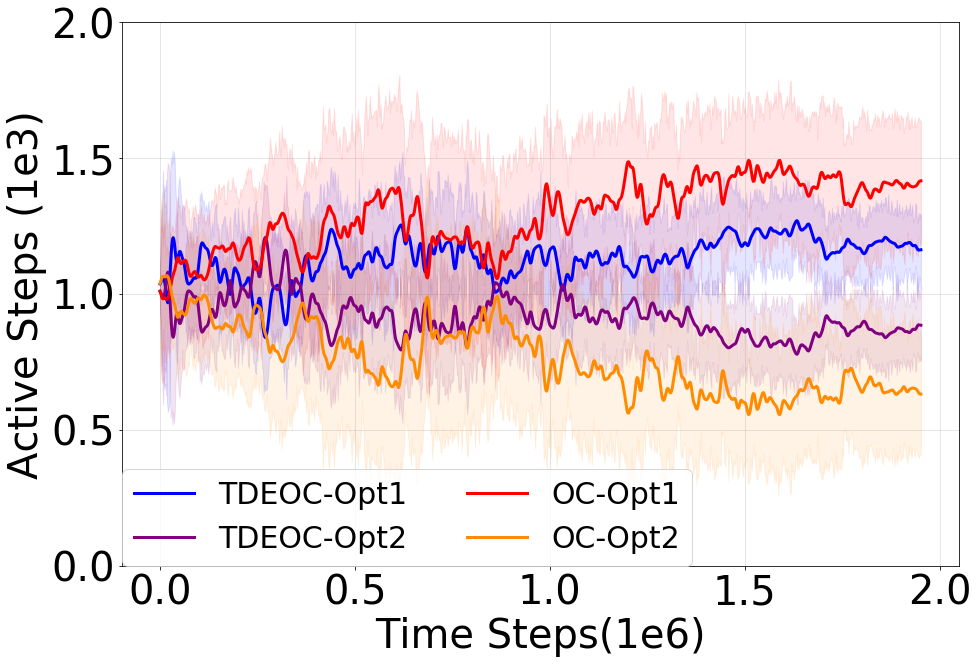}  \label{Fig_Tmaze_discrete_results}}
    \subfloat[OneRoom (Discrete)]{\includegraphics[scale=0.19]{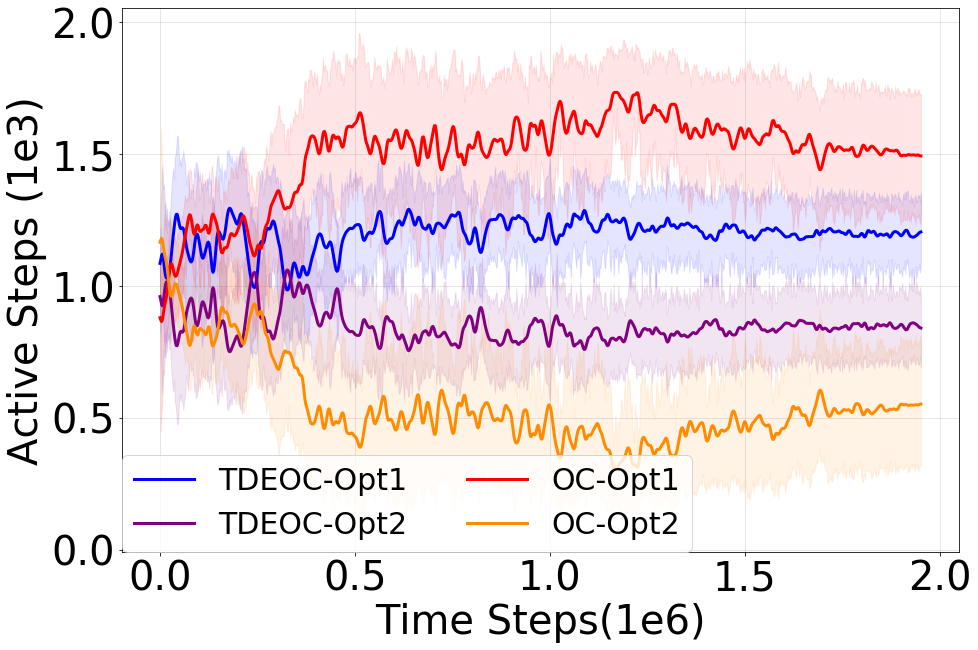}  \label{Fig_Oneroom_results}}\\
    \caption{\textbf{Plots represent the average activity of each option for TDEOC and OC}. The plots compare how the termination conditions used for TDEOC and OC contribute towards option activity. Mujoco and Miniworld experiments are recorded until two million steps  and ten million steps for Humanoid-v2. Each line represents an average over 20 independent runs.}
    \label{Fig_Option_Relevance}
\end{figure*}

\section{DEOC Results}
\begin{figure*}[!h]
    \centering
    \subfloat[HalfCheetah-v2]{\includegraphics[scale=0.12]{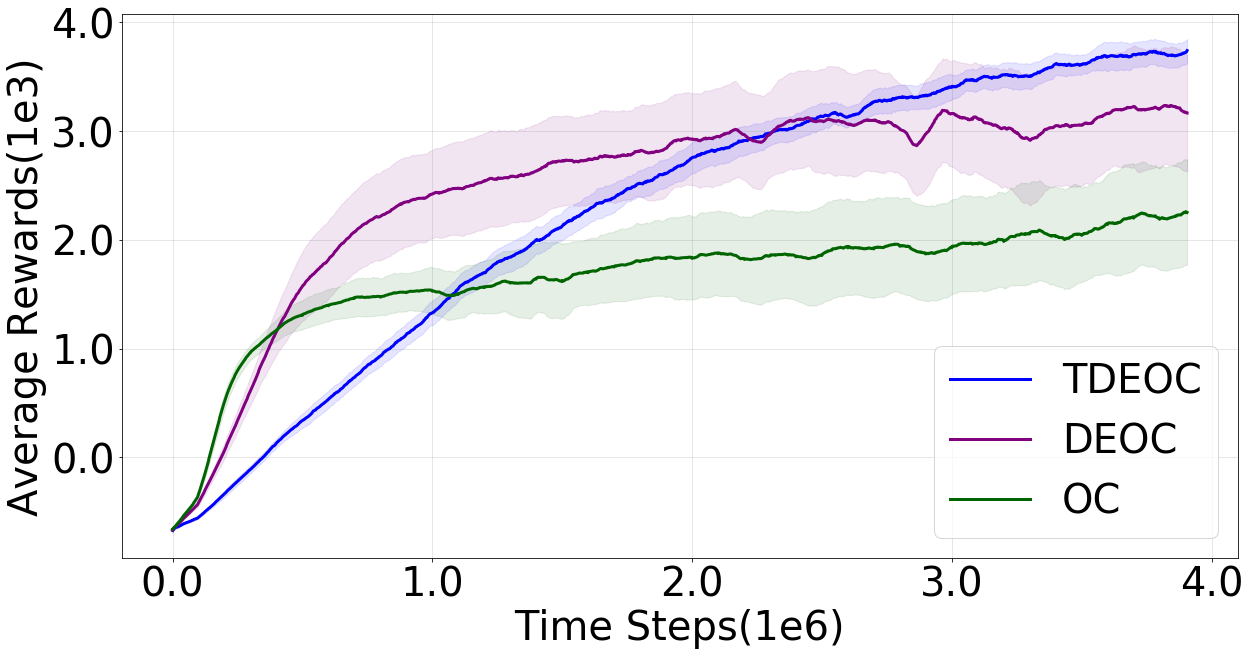}\label{Fig_DEOCvsTDEOC_HalfCheetah_results}}
    \subfloat[Hopper-v2]{\includegraphics[scale=0.12]{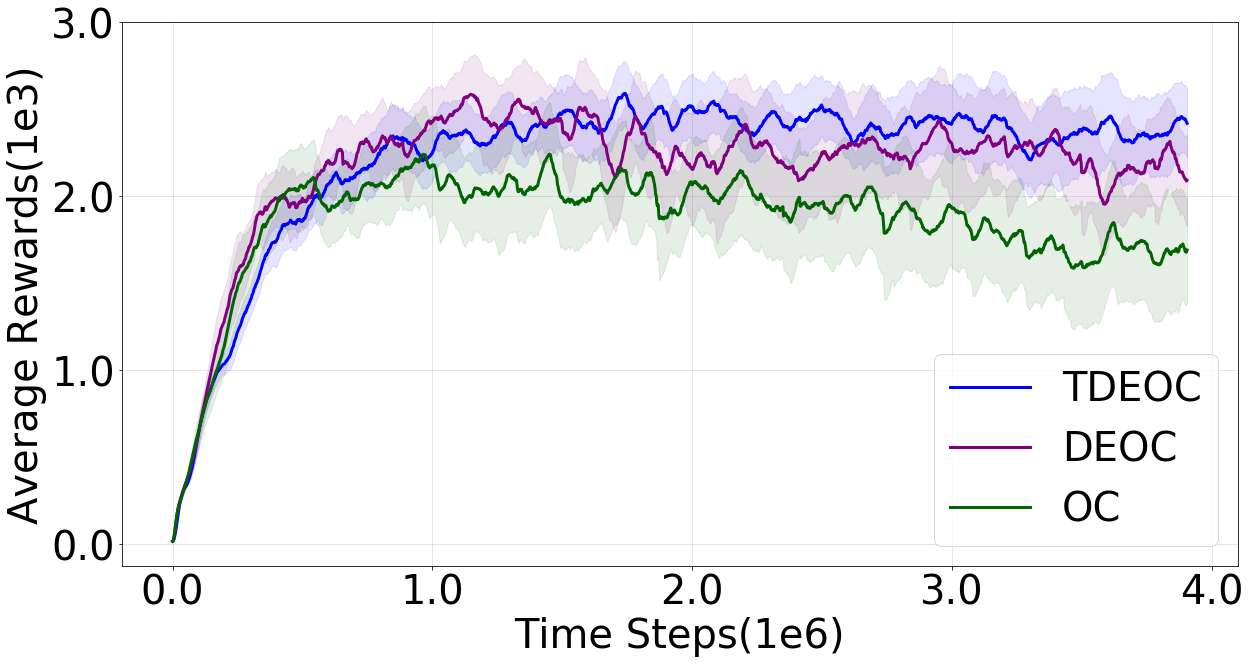}  \label{Fig_DEOCvsTDEOC_Hopper_results}} \\
    \subfloat[Walker2d-v2]{\includegraphics[scale=0.12]{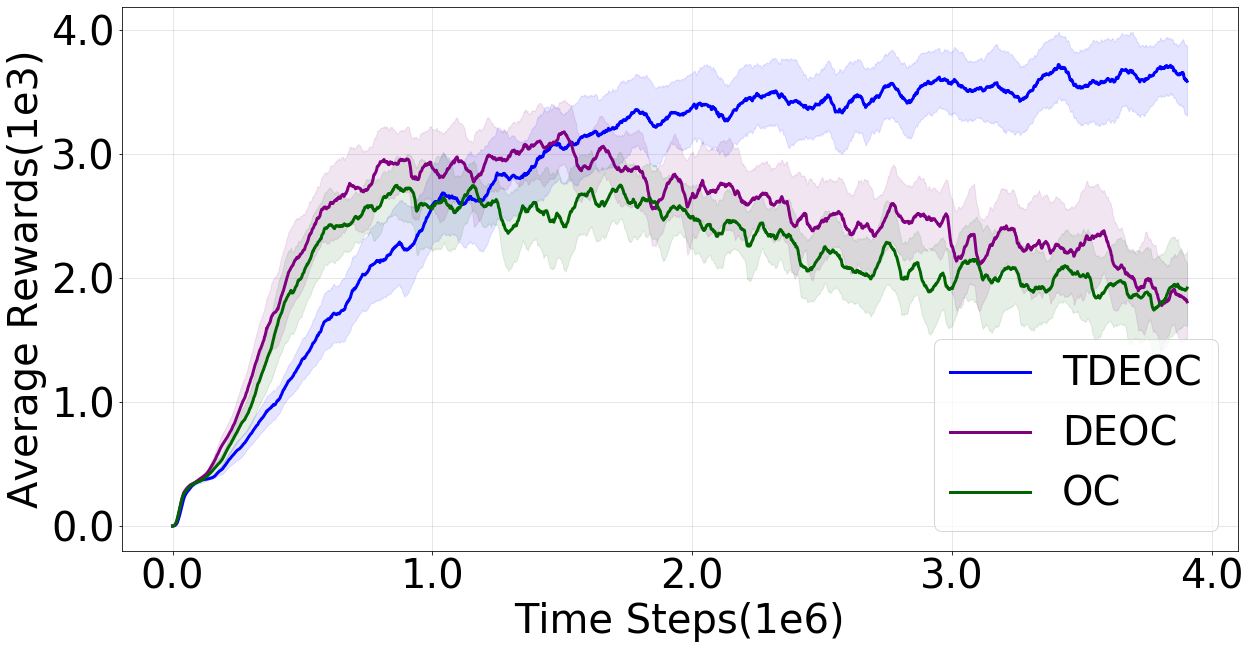}  \label{Fig_DEOCvsTDEOC_Walker_results}}\\
    \subfloat[Humanoid-v2]{\includegraphics[scale=0.12]{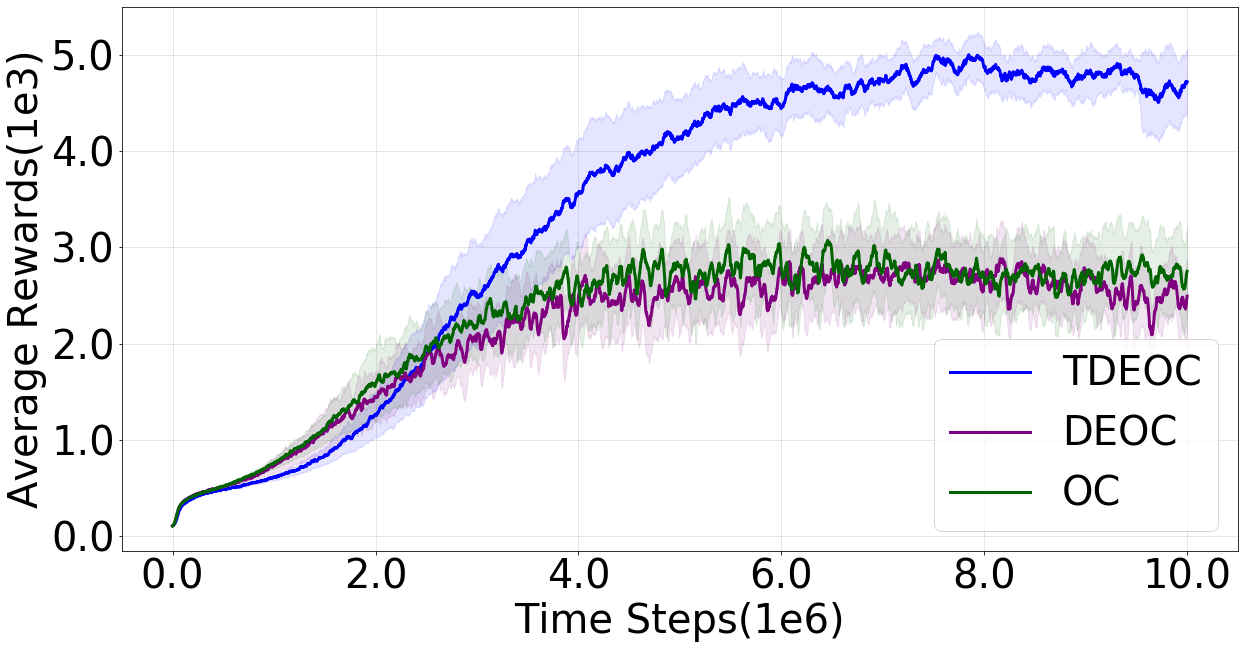} \label{Fig_DEOCvsTDEOC_Humanoid_results}} 
     \subfloat[Ant-v2]{\includegraphics[scale=0.12]{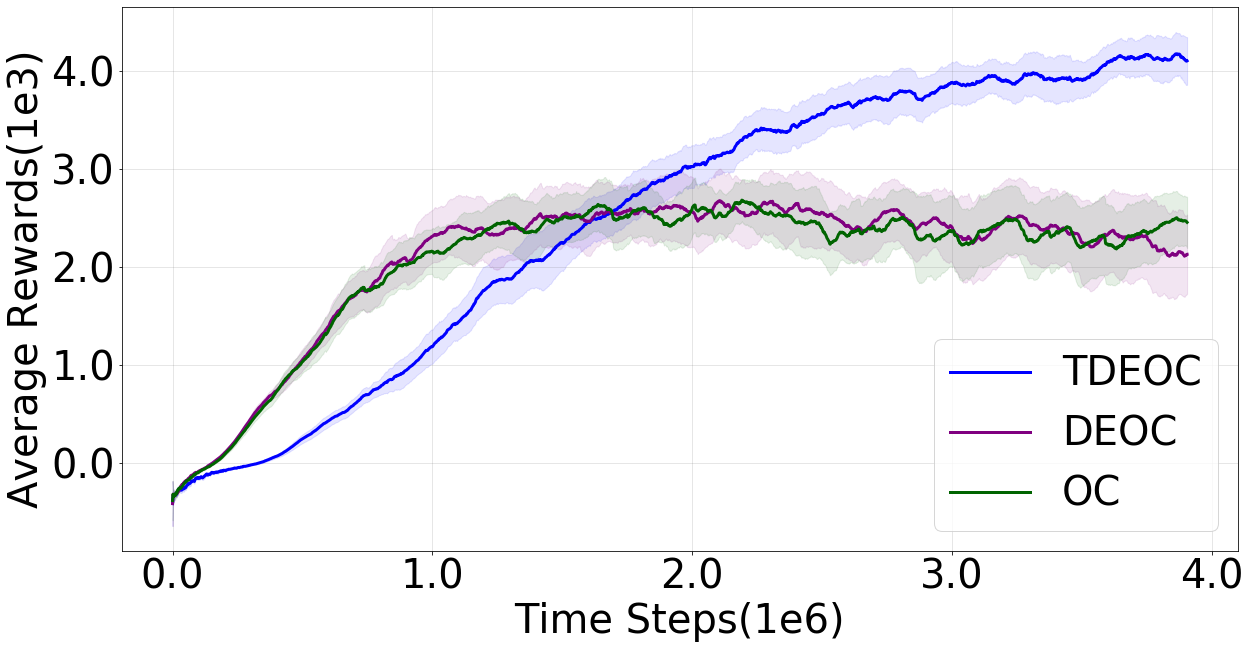} \label{Fig_DEOCvsTDEOC_Ant_results}}.
    \caption{\textbf{Empirical results showing DEOC runs along with TDEOC and OC}. Each line represents an average over 20 independent runs.}
    \label{Fig_DEOCvsTDEOC_results}
\end{figure*}

We present empirical results (Fig. \ref{Fig_DEOCvsTDEOC_results}) comparing DEOC against TDEOC and Option-Critic (OC) in this section. DEOC runs were omitted from Fig. \ref{Fig_TDEOC_results} for clarity and to retain emphasis on TDEOC's performance against option-critic.  While DEOC shows noticeable improvement in exploration and performance over OC, DEOC still suffers from degeneration as the updates are still the same as that of option-critic's. This is very observable in the Walker2d-v2 task (Fig. \ref{Fig_DEOCvsTDEOC_Walker_results}) where DEOC and OC suffer a drop in performance as one of the option dominates leading to a sub-optimal control by the least dominant option when selected due to noisy value or state estimates. TDEOC on the other hand generates useful and complementary options which explains its superior performance.

\section{Empirical Results for Hopper-v2 Task} \label{App_hopper}
In Section \ref{section_TDEOC_experiments}, we compare TDEOC against OC and PPO. We present the performance results for Hopper-v2 task from Mujoco which was absent in Fig \ref{Fig_TDEOC_results}. TDEOC not only outperforms OC and PPO, it is also capable of handling environment perturbations better.
\begin{figure}[!h]
    \centering
    \includegraphics[scale=0.2]{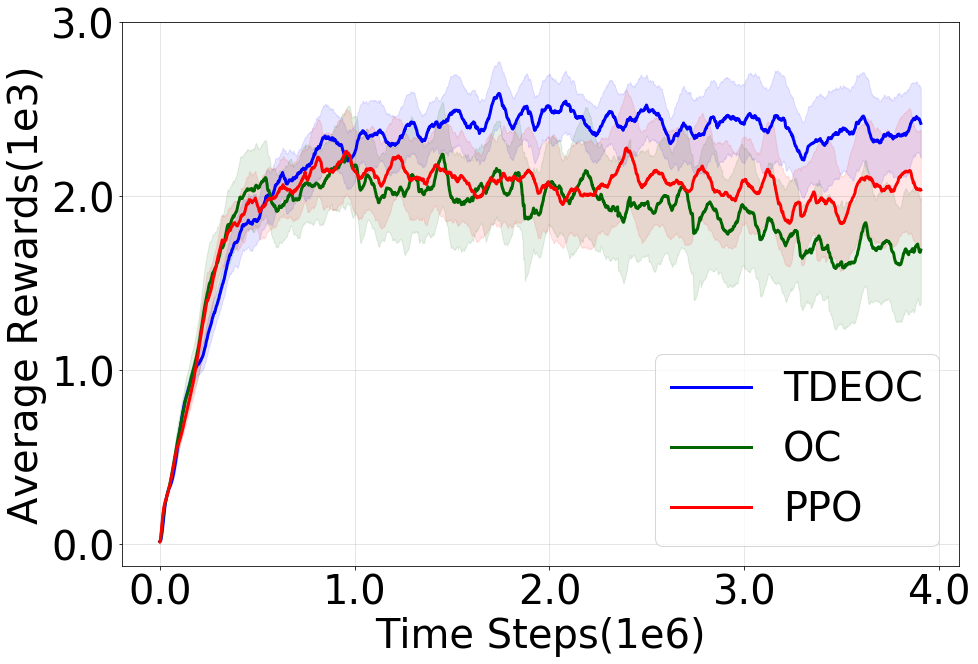}
    \caption{\textbf{Results on Hopper-v2 task from Mujoco.} }
    \label{fig:my_label}
\end{figure}
\section{Empirical Results for OneRoom Task} \label{App_oneroom}
In Section \ref{section_Interpreting_options} we visualize option trajectories for the OneRoom navigation task from Miniworld. The objective of the agent is to scan the room for the goal in the form of a red box located at a random location and navigate towards it. TDEOC manages to demonstrate comparable performance as PPO despite the added complexity of learning a hierarchy for a simple task.
\begin{figure}[!h]
    \centering
    \includegraphics[scale=0.2]{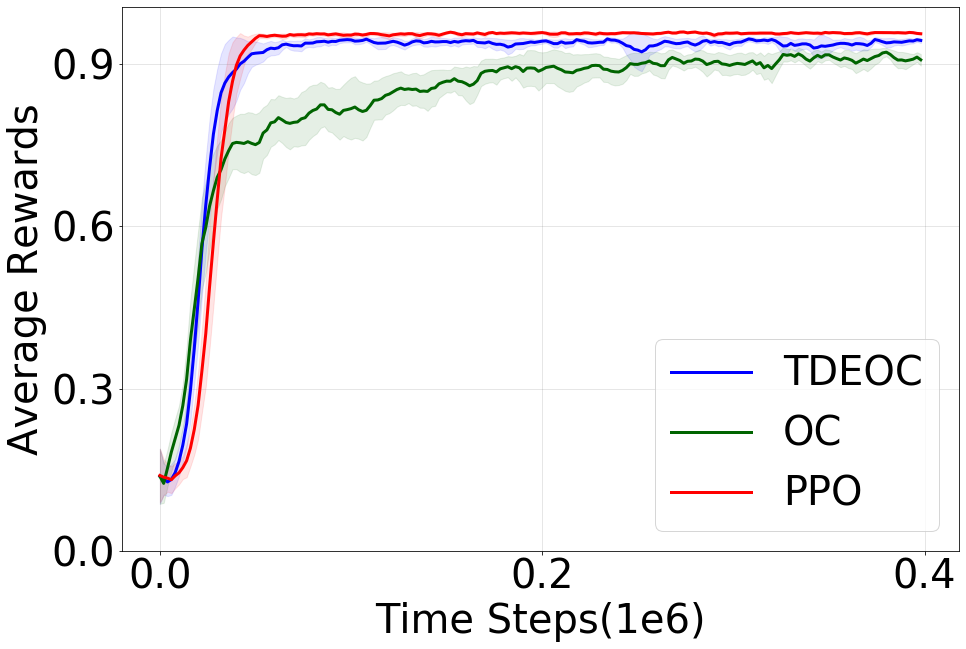}
    \caption{\textbf{Results on OneRoom task from Miniworld.} }
    \label{fig:my_label}
\end{figure}

\section{Reproducibility Checklist}
We list all the items of the Reproducibility Checklist \cite{reproducibility_checklist} and provide details for each point.

For all models and algorithms presented, check if you include:
\begin{itemize}
    \item \textbf{A clear description of the algorithm.} Please refer to Algorithm, \ref{deoc_algo}. We also provide the source code.
    
    \item \textbf{An analysis of the complexity (time, space, sample size) of any algorithm.} We do not perform this analysis.
    
    \item \textbf{A link to a downloadable source code, with specification of all dependencies, including external libraries.} We provide the source code along with our submission. We also explain the experimental setting in the paper and in the Appendix.
\end{itemize}

For any theoretical claim, check if you include:
\begin{itemize}
    \item \textbf{A statement of the result.} See main paper.
    \item \textbf{A clear explanation of any assumptions.} See main paper.
    \item \textbf{A complete proof of the claim. } See main paper.
\end{itemize}

For all figures and tables that present empirical results, check if you include:
\begin{itemize}
    \item \textbf{A complete description of the data collection process, including sample size.} For our tabular four-rooms task, we reuse the environment from \cite{bacon2017option}. We use standard versions of Mujoco \cite{todorov2012mujoco} and Miniworld \cite{gym_miniworld} for our single task evaluations. As for the transfer setting, we use custom environments using \cite{henderson2017multitask} for HalfCheetahWall-v0 and HopperIceWall-v0 and \cite{Khetarpal2020OptionsOI} for TMaze (continuous). We provide the source code along with all the environments used for our experiments.
    
    \item \textbf{A link to a downloadable version of the dataset or simulation environment.}
    See \cite{bacon2017option} for details regarding the four-rooms task and refer to our code for further details. Standard Mujoco \cite{todorov2012mujoco} and Miniworld \cite{gym_miniworld} tasks are available in their respective repositories. As for the custom tasks, please refer to our source code.
    
    \item \textbf{An explanation of any data that were excluded, description of any pre-processing step.}
    There is no need for any kind of pre-processing for the data collected from the environments. No data was excluded in any experiment we performed. 
    
    \item \textbf{An explanation of how samples were allocated for training / validation / testing.}
    As with all standard reinforcement learning control experiments, we do not split the data for training and validation. The performance during learning is presented in the main paper.
    
    \item \textbf{The range of hyper-parameters considered, method to select the best hyper-parameter configuration, and specification of all hyper-parameters used to generate results} For the tabular setting, we tune both (TDEOC and OC) for the best hyper-parameters for the given transfer setting. As for the standard Mujoco and Miniworld tasks, we reuse the hyper-parameters provided by baselines and \cite{gym_miniworld} for PPO. For OC we reuse the hyper-parameters provided by \cite{Klissarov2017LearningsOE}. As for the custom tasks, we tune the relevant hyper-parameters for each algorithm and compare the results based on the average results over 20 runs for each hyper-parameter configuration. All the final hyper-parameter configurations are reported in Appendix \ref{App_Implementation_details}.
    
    \item \textbf{The exact number of evaluation runs.} We average results over 300 runs for the tabular four-rooms case. All our experiments in the non-linear function approximation case are evaluated by averaging 20 independent seeds.
    
    \item \textbf{A description of how experiments were run.} Please refer to the experimental details provided in the main paper as well as the Appendix.
    
    \item \textbf{A clear definition of the specific measure or statistics used to report results.} In the four-rooms case, we use the number of steps to reach the goal as the performance measure for each algorithm. As for the non-linear function approximation case, we use average return to compare different algorithms. Our methods are consistent with standard performance measures used for the respective tasks.
    
    \item \textbf{Clearly defined error bars.} We plot error bars representing 0.5 of the standard deviation to observe differences in variance more acutely.
    
    \item \textbf{A description of results with central tendency (e.g. mean) \& variation (e.g. stddev ).} Our results are an average over all the runs. The error bars are half of the standard deviations for each run. 
    
    \item \textbf{A description of the computing infrastructure used.} A single CPU is used for the tabular as well as Mujoco based experiments. As for the Miniworld tasks, we use one CPU and a GPU for our results. 
    
\end{itemize}

\section{Implementation Details} \label{App_Implementation_details}

\subsection{Choice of Underlying algorithm} \label{App_choice_of_algo}
Our proposed termination objective relies on diversity estimates drawn from a well updated policy in order to identify the states where options tend to grow most diverse. This property is best observed in on-policy algorithms where samples from the same policy is used for policy evaluation and policy improvement. PPO is the state-of-the-art on-policy algorithm which is why it was used as the underlying algorithm. Although off-policy algorithms such as TD3 or SAC may achieve superior performance, these methods draw sample from a large buffer (mostly of capacity one million samples) which may even contain samples from the initial random policy, which dilutes TDEOC's ability to target most diverse states. Also, unlike PPO, TD3 and SAC have only been implemented for continuous control tasks while PPO has also been tested on discrete control tasks such as Miniworld \cite{gym_miniworld}. Another significant reason for choosing PPO is because option-critic had already been implemented using PPO updates (PPOC) \cite{baselines,Klissarov2017LearningsOE} and had been tested extensively, making our experiments standard, fair and reproducible. 

\subsection{Tabular Case} \label{APP_tabular}

For our four-rooms experiments, we reuse the implementations by \citet{bacon2017option}. Augmenting the reward with $ \mathcal{R}_{bonus} $ for tasks with very sparse rewards can cause the agent to prioritize diversifying options over learning the task. To mitigate this, we avoid the reward augmentation step in all our sparse reward tasks including the four-rooms task. Instead of standardizing the diversity values, use update a moving sum of all values observed in the current run and center the diversity around the moving mean instead. For more than three options, the diversity is computed by sampling six pairs of options and averaging the respective cross entropy. The sole reason of this is to avoid computing the mean of all samples at every step. Each algorithm is averaged over 300 runs. The code has been attached in the submission folder.
\begin{table}[hbt!]
    \centering
    \begin{tabular}{c|c|c}
    Hyper-parameter & TDEOC & OC \\
    \hline
     Termination lr & 5e-2 & 1e-1\\
     Intra-Option lr & 1e-2 & 1e-2 \\
     Critic lr & 5e-1 & 5e-1 \\
     Action Critic lr & 5e-1 & 5e-1 \\
     Discount & 0.99 & 0.99 \\
     Max Steps & 1000 & 1000 \\
     No of Options & 4 & 4 \\
     Temperature & 1e-3 & 1e-3 \\

     \end{tabular}
    \caption{Hyper-parameters for Tabular Four-rooms task}
    \label{tab:my_label}
\end{table}

\subsection{Non-Linear Function Approximation Case} \label{App_Nonlinearcase}
We provide the implementation details as well as the hyper-parameters for all our non-linear function approximation cases. The hyper-parameters for PPO and OC are consistent with those suggested in \citet{baselines} and \citet{Klissarov2017LearningsOE} respectively. We use two critics for our algorithms (DEOC and TDEOC) as used by \citet{bacon2017option} in their tabular case implementation. All our plots are averaged over 20 independent runs with the error bounds representing 0.5 of the standard deviation. 

\subsubsection{}
For standard Mujoco tasks, we incorporate our algorithm within the PPOC code \cite{Klissarov2017LearningsOE}. The pseudo reward bonus is scaled down depending on the task with the intention of prioritizing task reward. The diversity term is calculated using cross entropy, as stated in the Eq. \eqref{eq_pseudoreward}. We compute the softmax of the action distribution before computing the cross entropy to ensure $ \mathcal{R}_{bonus} $ remains positive. As mentioned earlier, for the TMaze(continuous) task, we avoid augmenting the reward with the diversity term. As for the HalfCheetahWall-v0 and HopperIce-v0, we reuse the resources provided by \cite{henderson2017multitask}. The obstacle is observed in the agent's state space only when the agent is within one metre of the obstacle. The learning rate for TDEOC is slightly slower than OC to allow the algorithm sufficient time to learn the bottleneck states. We average runs over 20 sequential seeds starting from seed 10 for our results.

\begin{table}[hbt!]
    \centering
    \begin{tabular}{c|c}
    Hyper-parameter & Value \\
    \hline
     Termination lr & 5e-7 \\
     Termination lr TMaze(Continuous) & 5e-8 \\
     Timesteps per batch & 2048 \\
     Optim epochs & 10 \\
     Clip param & 0.2 \\
     Entropy coefficient & 0.0 \\
     Gamma & 0.99 \\
     Lambda & 0.95 \\
     Lr schedule & constant
    \end{tabular}
    \caption{Common hyper-parameters across all continuous control tasks}
    \label{tab:my_label}
\end{table}

\begin{table}[hbt!]
    \centering
    \begin{tabular}{c|c|c|c}
    
    Environment & TDEOC & DEOC & OC \\
    \hline
    Ant-v2 & 1e-4 & 3e-4 & 3e-4 \\
    HalfCheetah-v2 & 1e-4 & 3e-4 & 3e-4 \\
    Hopper-v2 & 1e-4 & 3e-4 & 3e-4 \\
    Walker2d-v2 & 1e-4 & 3e-4 & 3e-4 \\
    Humanoid-v2 & 3.33e-5 & 1e-4 & 1e-4 \\
    HalfCheetahWall-v0 & 1e-4 & nan & 1e-4 \\
    HopperIceWall-v0 & 1e-4 & nan & 1e-4 \\
    TMaze (Continuous) & 3e-5 & nan  & 1e-4 \\

    \end{tabular}
    \caption{Learning rates for various continuous control tasks}
    \label{tab:my_label}
\end{table}

\begin{table}[hbt!]
    \centering
    \begin{tabular}{c|c}
    
    Environment & Trade-off\\
    \hline
    Ant-v2 & 0.2 \\
    HalfCheetah-v2 & 0.7  \\
    Hopper-v2 & 0.2 \\
    Walker2d-v2 & 0.2 \\
    HalfCheetahWall-v0 & 0.6 \\
    HopperIceWall-v0 & 0.4 \\
    TMaze (Continuous) & 0.0 \\

    \end{tabular}
    \caption{Trade-off value for various control tasks}
    \label{tab:my_label}
\end{table}

\subsubsection{}
For Miniworld tasks, we use the code provided by baselines \cite{baselines} for Atari environments and implement the PPOC networks consistent with \cite{Klissarov2017LearningsOE} within. The objective of the Sidewalk task is for the agent to navigate to a goal object placed at the end of the street. The episode terminates and the environment resets if the agent strays away from the path onto the street. The OneRoom task can be perceived as a simpler version of the Sidewalk task, solely consisting of navigating to the goal object placed randomly anywhere in the room. As for the TMaze environment, the objective is to navigate to a goal object placed randomly at either ends of the horizontal bottom hallway. Due to discrete action space, diversity is computed by taking the softmax of the logits of the policy network. Rest of the implementation is consistent with the continuous control case discussed above. Due to the sparse reward situation, we again do not augment the reward with the diversity term. Like in the continuous control case, we average runs over 20 sequential seeds starting from seed 10 for our results. The hyper parameters used are given below:
\begin{table}[!h]
    \centering
    \begin{tabular}{c|c}
    Hyper-parameter & Value \\
    \hline
     Termination lr & 5e-7 \\
     Timesteps per batch & 2048 \\
     Optim epochs & 4 \\
     Entropy coefficient & 0.1 \\
     Clip Param & 0.2 \\
     Gamma & 0.99 \\
     Lambda & 0.95 \\
     Lr schedule & linear
    \end{tabular}
    \caption{Common hyper-parameters across all Miniworld tasks}
    \label{tab:my_label}
\end{table}
\begin{table}[!h]
    \centering
    \begin{tabular}{c|c|c}
    
    Environment & TDEOC & OC \\
    \hline
    MiniWorld-OneRoom-v0 & 1e-4 & 3e-4 \\
    MiniWorld-Sidewalk-v0 & 1e-4 & 3e-4 \\
    MiniWorld-TMaze-v0 & 1e-4 & 3e-4  \\
    \end{tabular}
    \caption{Learning rates for various Miniworld tasks}
    \label{tab:my_label}
\end{table}

\end{document}